\pdfoutput=1
\documentclass[11pt]{article}
\usepackage{lipsum}

\usepackage{amsmath}
\usepackage{booktabs}
\usepackage[final]{acl}
\usepackage{graphicx} %插入图片的宏包
\usepackage{float} %设置图片浮动位置的宏包
\usepackage{subfigure} %插入多图时用子图显示的宏包
\usepackage{times}
\usepackage{latexsym}
\usepackage{bbding}
\usepackage{ulem}
\usepackage[marginal]{footmisc}

\usepackage{cleveref}
\crefname{section}{§}{§§}
\Crefname{section}{§}{§§}

\usepackage{listings}
\usepackage{xcolor}
\usepackage{pifont}
\newcommand{\whiteding}[1]{\ding{\numexpr171+#1\relax}}

\usepackage[T1]{fontenc}
\usepackage[utf8]{inputenc}

\usepackage{microtype}
\usepackage{lipsum} % 用于生成示例文本
\usepackage{pagenote} % 用于在不同页面插入脚注
\usepackage{inconsolata}

\usepackage{inconsolata}
\usepackage{longtable}
\usepackage{hyperref}

\usepackage{xcolor}

\usepackage{tabularx}

\usepackage{inconsolata}
\usepackage{graphicx}
\usepackage{inconsolata}
\usepackage{array}
\usepackage{mathrsfs}
\usepackage{amsthm}
\usepackage{amsmath}
\usepackage{amsfonts}

\usepackage[noend]{algpseudocode}
\usepackage{multirow}
\usepackage{booktabs}
\usepackage{color}
\usepackage{arydshln}
\usepackage{makecell}
\usepackage{soul}
\usepackage{cleveref}
\crefname{section}{§}{§§}
\Crefname{section}{§}{§§}

\title{C\textsc{heck}W\textsc{hy}: Causal Fact Verification via Argument Structure}

\author{Jiasheng Si\textsuperscript{\rm 1,4$\dagger$}, Yibo Zhao\textsuperscript{\rm 2,3$\dagger$}, Yingjie Zhu\textsuperscript{\rm 2,3}, Haiyang Zhu\textsuperscript{\rm 2,3}, Wenpeng Lu\textsuperscript{\rm 1,4}, Deyu Zhou\textsuperscript{\rm 2,3*} \\
     \textsuperscript{\rm 1}Key Laboratory of Computing Power Network and Information Security,  Ministry of Education, \\ Shandong Computer Science Center, Qilu University of Technology (Shandong Academy of
Sciences), China\\
\textsuperscript{\rm 2} School of Computer Science and Engineering, Southeast University, China \\
\textsuperscript{\rm 3} Key Laboratory of New Generation Artificial Intelligence Technology and Its 
Interdisciplinary \\ Applications (Southeast University),  Ministry of Education, China \\
\textsuperscript{\rm 4} Shandong Provincial Key Laboratory of Computer Networks, Shandong Fundamental \\ Research Center for Computer Science, China \\
    \texttt{\{jiashengsi, lwp\}@qlu.edu.cn}, \texttt{\{yibozhao, yj\_zhu, haiyangzhu, d.zhou\}@seu.edu.cn} \\
    }

\begin{document}
\maketitle
\newcommand\blfootnote[1]{%
	\begingroup
	\renewcommand\thefootnote{}\footnote{#1}%
	\addtocounter{footnote}{-1}%
	\endgroup
}

\blfootnote{\textsuperscript{$\dagger$} Equal Contribution.}
\blfootnote{\textsuperscript{*} Corresponding Author.}

\begin{abstract}
	With the growing complexity of fact verification tasks, the concern with ``thoughtful'' reasoning capabilities is increasing.
	However,
	recent fact verification benchmarks mainly focus on checking a narrow scope of semantic factoids within claims and lack an explicit logical reasoning process.
	In this paper, we introduce \textsc{CheckWhy}, a challenging dataset tailored to a novel causal fact verification task: checking the truthfulness of the causal relation within claims through rigorous reasoning steps.
	\textsc{CheckWhy} consists of over 19K ``\textit{why}'' \textit{claim-evidence-argument structure} triplets with \textit{supports}, \textit{refutes}, and \textit{not enough info} labels. Each argument structure is composed of connected evidence, representing the reasoning process that begins with foundational evidence and progresses toward claim establishment. Through extensive experiments on state-of-the-art models, we validate the importance of incorporating the argument structure for causal fact verification. Moreover, the automated and human evaluation of argument structure generation reveals the difficulty in producing satisfying argument structure by fine-tuned models or Chain-of-Thought prompted LLMs, leaving considerable room for future improvements\footnote{The dataset and code are available at \url{https://github.com/jasenchn/checkwhy}}.

\end{abstract}

\section{Introduction}

Fact verification is a crucial debunking task that entails verifying the truthfulness of claims by cross-referencing them with reliable evidence drawn from established resources~\cite{DBLP:journals/tacl/GuoSV22},
which prevents the proliferation of erroneous information online and fosters public trust~\cite{lewandowsky2020debunking, DBLP:conf/emnlp/Glockner0G22}.
However, with the multi-step reasoning capability in fact verification models remaining uncertain~\cite{DBLP:conf/emnlp/SchusterSYFSB19,DBLP:conf/emnlp/PanLKN23,causalwalk},
existing research reflects the deficiency in in-depth understanding of the explicit reasoning mechanisms when performing inference on multi-hop evidence.
This prompts us to develop a strong benchmark that incorporates the interpretable ``thought'' process to assess the logical reasoning capabilities of models.

Currently, substantial progress has been made on common fact verification benchmarks, e.g., H\textsc{o}V\textsc{er}~\cite{jiang2020hover}, FEVEROUS~\cite{aly2021feverous}, and S\textsc{ci}T\textsc{ab}~\cite{DBLP:conf/emnlp/LuPLNK23}.
Nevertheless, existing resources have inherent limitations.
Classic datasets primarily focus on verifying the semantic factoids of ``\textit{who}'', ``\textit{what}'', ``\textit{when}'', and ``\textit{where}''  within the claim~\cite{rani2023factify}.
For example, verifying the claim ``\textit{John Lennon was born before the astronaut who drank the first coffee in space.}'' can be decomposed into verifying factoids such as ``\textit{who, and when the astronaut drank the first coffee in space.}'' and ``\textit{when the man was born.}''.
However, these semantic factoids can be answered individually by straightforward factoids-matching between the claim and distinct independent evidence~\cite{jiang2020hover,DBLP:conf/emnlp/PanLKN23},
e.g., word overlapping or proof matching\cite{DBLP:journals/tacl/Krishna0022},
thereby limiting its significant potential to summarize the correlational evidence with ``thought'' steps.
Heretofore, noticeably absent in prior datasets are ``\textit{why}'' claims: containing causal relations that need to be verified. These claims prompt for not simple factoid matching, but an explicit logical reasoning path for verification~\cite{ho2022wikiwhy}.

\begin{figure*}[htbp]
	\centering
	\includegraphics[width=12cm]{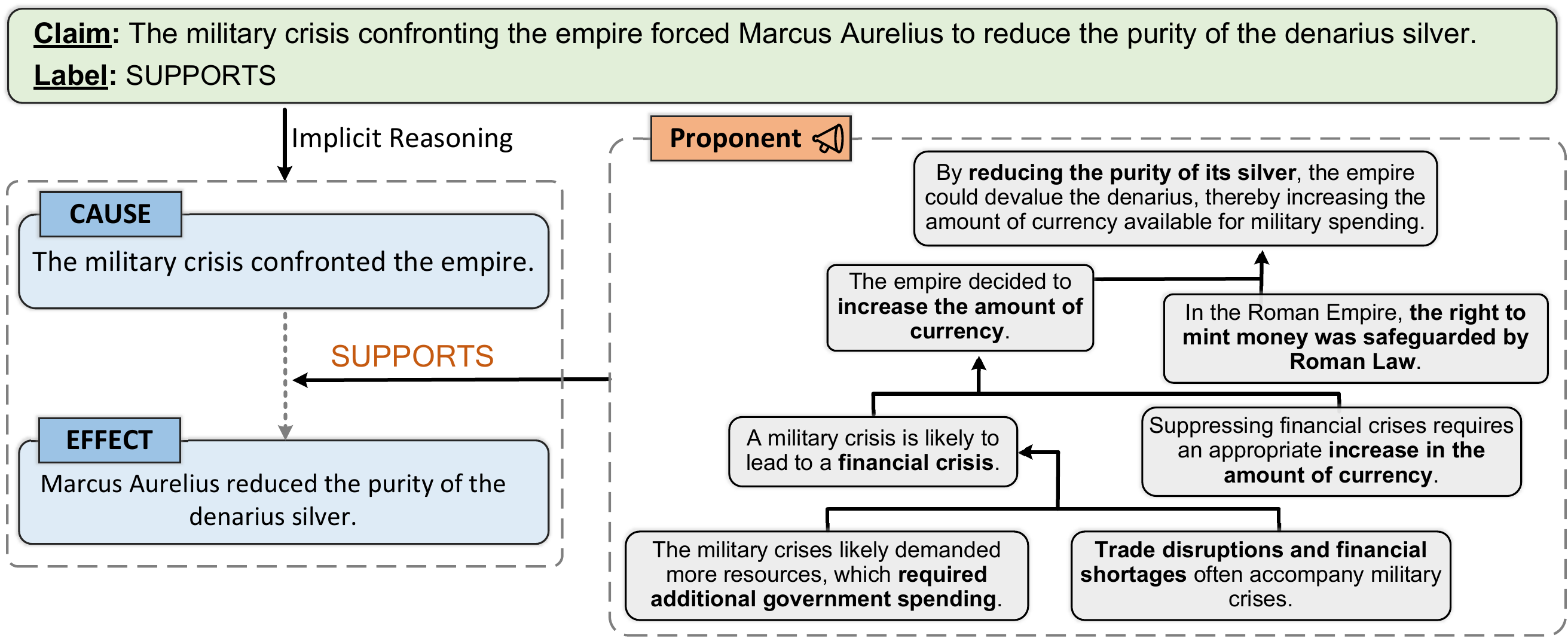}
	\vspace{-1ex}
	\caption{An entry from C\textsc{heck}W\textsc{hy}: a ``\textit{why}'' \textbf{claim} with its corresponding \textbf{cause} and \textbf{effect}, and an \textbf{argument structure} representing the reasoning process from cause to effect. Notably, the cause-effect pair is used solely during the annotation process and not included in the argument structure, implying that it is implicitly inferred from the claim, rather than being provided explicitly.}
	\label{argument structure example}
\end{figure*}

More specifically,
Figure~\ref{argument structure example} presents a ``\textit{why}'' claim featuring a cause-effect pair where ``\textit{military crises}'' causes the ``\textit{decrement of the purity of denarius silver.}''.
Verifying such causal relations is quite challenging,
which necessitates complex logical reasoning and context information beyond the factoids within the claim~\cite{NEURIPS2023_631bb943,jin2023can,romanou2023crab}.
For instance,
to \textit{support} this causal relations (i.e., \textit{military crises} $\xrightarrow[]{cause}$ \textit{decrease purity of silver}),
it is essential to incorporate the extra intermediate reasoning steps to bridge the connection between the cause-effect pair,
thus forming a logical reasoning path: \textit{military crises} $\xrightarrow[]{\text{\whiteding{1}}}$ \textit{financial crisis} $\xrightarrow[]{\text{\whiteding{2}}}$ \textit{increase amount of currency} $\xrightarrow[]{\text{\whiteding{3}}}$ \textit{decrease purity of silver}.
The rationale behind \whiteding{1} is that \textit{military crises} often coincide with extra factors such as \textit{trade disruption} or \textit{more government spending}, leading to the \textit{financial crisis}.
The reason supporting \whiteding{2} is that \textit{increase amount of currency} is a demanding response to suppress the \textit{financial crisis}.
Furthermore,
\textit{the rights to mint money that is safeguarded by Roman Law} ensures the behavior of \whiteding{3} is established.
The above inference reveals a coherent reasoning process,
which involves aligning the construction of a ``thoughtful'' logical structure among correlational evidence with causal relation verification.

In this paper,
we introduce C\textsc{heck}W\textsc{hy}, a challenging dataset built around a novel causal fact verification task: assessing whether the causal relation within the claim is valid by explicit logical structure.
This dataset consists of 19,596 \textit{claim-evidence-argument structure} triplets derived from the \textsc{WikiWhy} dataset~\cite{ho2022wikiwhy}.
The uniqueness of C\textsc{heck}W\textsc{hy} is that each entry contains a ``\textit{why}'' \textit{claim} with causal relations and an \textit{argument structure} formed by correlated evidence:
the latter is inspired by the theory literature on argument structure~\cite{grimshaw1990argument, freeman2011argument},
which depicts how the different statements fit together as wholes to allegedly lend support to the claim.
Moreover, inspired by prior research~\cite{Glockner2021AmbiFCFA}, we assume that the label of a causal relation within the claim depends on the provided argument structures, rather than the semantics itself. Thus, each claim is labeled as \textit{supports}, \textit{refutes}, or \textit{not enough info} based on different argument structures.
In addition,
to prevent the bias in human cognition,
we employ a human-model collaboration annotation approach
to generate claims, evidence, and corresponding argument structures.
Compared to existing datasets,
\textsc{CheckWhy} covers a variety of topics and argument structures,
which may prove valuable for performing causal reasoning across various scenarios.

\iffalse
	We propose a new causal fact verification task to assess whether the causal relation described in the claim is valid via logical structure and introduce the corresponding C\textsc{heck}W\textsc{hy} dataset, which consists of a total of 19,596 samples derived from the \textsc{WikiWhy} dataset~\cite{ho2022wikiwhy}.
	The uniqueness of C\textsc{heck}W\textsc{hy} lies in the fact that each entry contains a ``\textit{why}'' \textit{claim} with causal relations and an \textit{argument structure} formed by correlated evidence. The latter is inspired by the theory literature on argument macrostructure~\cite{grimshaw1990argument, freeman2011argument},
	which depicts how different statements fit together as wholes to allegedly lend support to the claim.
	We adopt a complex annotation strategy in which each claim is labeled as \textit{supports}, \textit{refutes}, or \textit{not enough info} based on different argument structures. We assume that the label of the causal relation within the claim depends on the provided argument structures, rather than the semantics itself.
	Moreover,
	to prevent the bias in human cognition,
	we employ a human-model collaboration annotation approach,
	as depicted in Figure~\ref{Generate-edit-filter},
	to generate claims, evidence, and corresponding argument structure.
	Compared to existing datasets,
	\textsc{CheckWhy} covers a variety of topics and argument structures,
	which may prove valuable for developing the skill of causal reasoning across various scenarios.\
\fi

Based on the experiments on four tasks we propose and the human evaluation in our \textsc{CheckWhy},
our experiments reveal the significance of incorporating the argument structure for causal fact verification.
Meanwhile,
our experiments in argument structure generation also validate the difficulty in producing satisfying argument structures for causal claims.
Our key contributions are summarized as follows:
(I) We propose verifying the ``why'' claims with causal relations through reasoning on argument structure as a novel causal fact verification formulation.
(II) We construct \textsc{CheckWhy} by introducing a human-model collaboration annotation approach,
drawing inspiration from the theory research on argument structure.
(III) We conduct thorough experiments on state-of-the-art models with four tasks, including fine-tuned models and LLMs,
which investigates various settings and points out the potential for improvement.

\section{Preliminaries}

Our \textsc{CheckWhy} framework is inspired by the argument structures theorized in the literature on logical theory~\cite{thomas1973practical,toulmin2003uses,walton2008argumentation,walton2013argumentation}.
Thus, we begin by outlining the standard argumentation structure and then present a concise overview of our argument structure.

\begin{figure*}[htbp]
	\centering
	\includegraphics[width=0.63\textwidth]{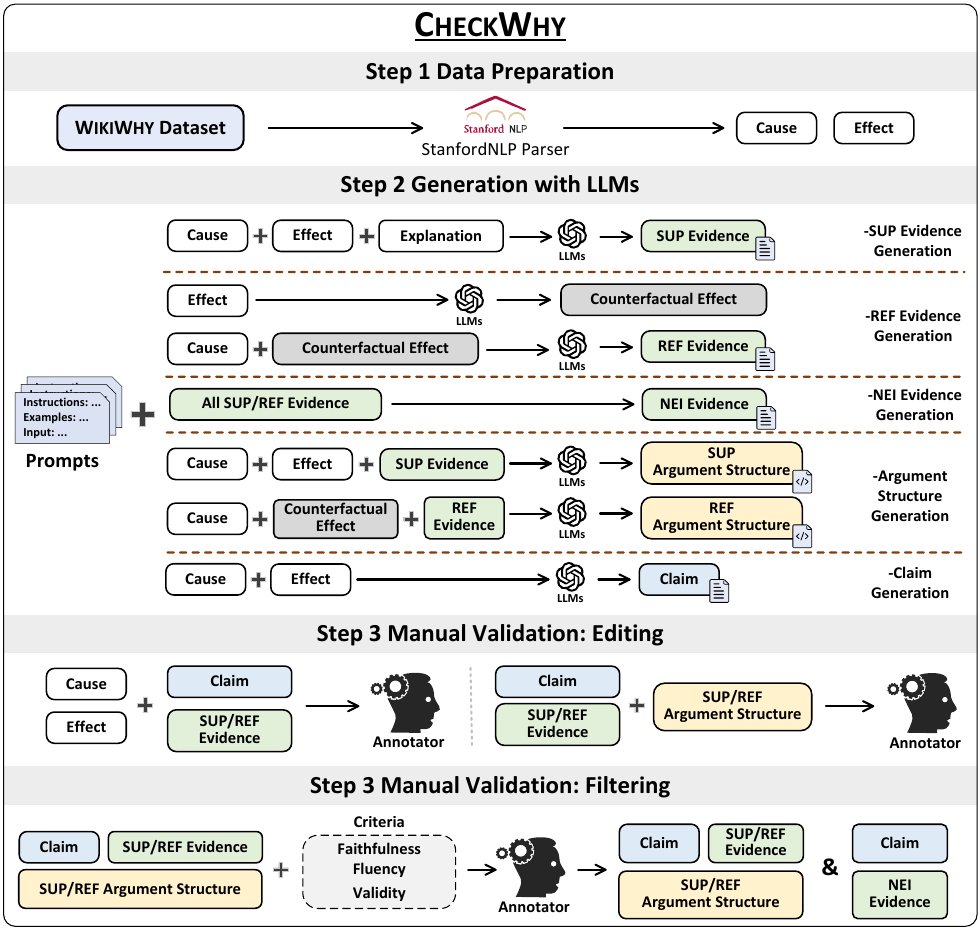}
	\vspace{-1ex}
	\caption{The human-model collaboration annotation process of \textsc{CheckWhy}, which contains three steps:
		(I) data preparation; (II) generation with LLMs (including the generation of claim, evidence, and argument structure); (III) manual validation.}
	\label{Generate-edit-filter}
\end{figure*}

\subsection{Standard Argumentation Structure}
\label{sec: Structure of Argumentation}
The standard argumentation structure \cite{thomas1973practical} is the scheme for structurally representing argument macrostructure: concerning what are the structural patterns in which the elements that constitute an argument may combine to form the overall argument.
It consists of four basic structures.

\noindent $\bullet$ \textbf{Serial Argument ($P\rightarrow C$)}:
an argument structure has one premise $P$ to give a reason to support the conclusion $C$.

\noindent $\bullet$ \textbf{Convergent Argument ($P_1 \vee P_2\rightarrow C$)}:
an argument structure has more than one premise, with each one function separately ($P_1\vee P_2$) as a reason to support the conclusion $C$.

\noindent $\bullet$ \textbf{Linked Argument ($P_1 \land P_2\rightarrow C$)}:
an argument structure has more than one premise, and the premises function together ($P_1\land P_2$) to give a reason to support the conclusion $C$.

\noindent $\bullet$ \textbf{Divergent Argument ($P\rightarrow C_1/C_2$)}:
an argument structure has more than one separate conclusion ($C_1/C_2$) that can be supported by the same premise $P$.

\subsection{Our Argument Structure}
\label{subsec: our argument structure}
The argument structure in \textsc{CheckWhy} follows a tree-like framework,
with the claim serving as the root node and evidence branching out as child nodes.
These nodes are connected by directed edges that symbolize logical relations.
Specifically, following the standard argumentation structure~\cite{thomas1973practical}, we blend various basic structures into a unified argument tree. This is achieved by referring to the semantics of each piece of evidence and the logical relations between different pieces of evidence. In this structure, the \textbf{series argument} leads from one child node to one parent node, while the \textbf{divergent argument} deduces multiple parent nodes from one child node.
Furthermore,
due to the challenging to discern the subtle distinction between \textit{convergent} and \textit{linked argument},
we merge these structural types into a new \textbf{combined argument}.
This argument tree formalizes the reasoning process that begins with foundational evidence and progresses toward claim establishment.

A tricky issue here is that the argument structure inherently \textit{supports} in the claim based on its definition,
whereas the \textit{refutes} instances are indispensable for causal fact verification.
To tackle this,
drawing inspiration from the \textit{warrant} and \textit{rebuttal} concepts in Toulmin's structure~\cite{toulmin2003uses},
wherein the warrant offers facts or rules to back up the claim, and the rebuttal indicates conditions that negate the claim.
We apply diverse argument structures to the same claim to collect both \textit{supports} and \textit{refutes} instances.
In specific,
offering the basic argument structure (i.e., \textit{warrant}) to acquire \textit{support} labels, and providing another structure (i.e., \textit{rebuttal}) that upholds the opposition of causal relation in the claim to obtain \textit{refutes} labels.

\section{The \textsc{CheckWhy} Dataset}

We adopt a human-model collaboration annotation approach to construct our \textsc{CheckWhy},
as shown in Figure~\ref{Generate-edit-filter},
which including data preparation (\cref{subsec.data_preparation}),
generation with LLMs (\cref{subsec.Generation_with_LLMs}),
and manual validation (\cref{subsec.manual_validation}).

\subsection{Data Preparation}
\label{subsec.data_preparation}
We utilize the \textsc{WikiWhy} dataset \citep{ho2022wikiwhy}, a publicly available cause-and-effect QA resource, as our data source.
Each QA pair in this dataset was annotated with effect (question), cause (answer), and associated explanations,
making it well-suited for our purposes.
From the entire collections of cause-effect pairs in \textsc{WikiWhy},
we initially filter out instances that contain incomplete syntactic structure in either the cause or the effect,
thus remaining cause-effect pairs with valid causal relations.
This is realized by justifying the absence of a \textit{predicate component} in the sentence using StanfordNLP Parser tool \citep{qi2020stanza}.
To ensure quality, we include an option in the subsequent manual validation process to mark a claim as "Discard - not a valid causal relation, incomplete, or redundant". During this stage, we extract 7,403 cause-effect pairs from a total of 9,406 instances within the \textsc{WikiWhy} dataset.

\subsection{Generation with LLMs}
\label{subsec.Generation_with_LLMs}
The rich knowledge and generative capabilities of large language models (LLMs) raise widespread attention on their use in aiding the construction of dataset \citep{si2023large, chen-etal-2023-disco}.
As such, we adopt GPT-4 \citep{achiam2023gpt} for our data generation through in-context learning \cite{ouyang2022training}.
Notably,
for each cause-effect pair,
our \textsc{CheckWhy} applies three argument structures to each claim to acquire the \textit{SUP}, \textit{REF}, and \textit{NEI} instances with \textit{supports}, \textit{refutes}, and \textit{not enough info} labels, which ensures a dataset with balanced distribution of veracity labels. All the prompts used in our annotation process are presented in Appendix ~\ref{Generation Prompts}.

$\bullet$ \textbf{Support Evidence Generation} \quad
To obtain valid \textit{SUP} instances,
textual evidence that \textit{support} the claim is generated via GPT-4.
In specific,
by taking the cause-effect pair and associated explanation within \textsc{WikiWhy} as input,
the GPT-4 is prompted with the designed \texttt{ReAct}-like~\cite{yao2022react} examples,
enabling LLMs to logically deduce the effect from the cause through a step-by-step thought process.
Finally,
the reasoning process generated by LLMs is considered as \textit{support} evidence.

$\bullet$ \textbf{Refute Evidence Generation} \quad
Due to the significantly broader generation scope compared to the \textit{SUP} instances,
building \textit{REF} instances is a challenging task \cite{zhu2023explain,tan2023multi2claim}.
To produce valid evidence that refutes the claim, following \citet{zhu2023explain},
we apply the generation of evidence that refutes the counterfactual effect as an intermediate step.
In specific, starting with the original cause, we prompt GPT-4 to generate the counterfactual effect by crafting an effect with an opposite label.
Then,
we prompt GPT-4 with \texttt{ReAct}-like examples to reason from cause to the counterfactual effect,
resulting in the generation of evidence that refutes the original effect.

$\bullet$ \textbf{NEI Evidence Generation} \quad
Following \citet{aly2021feverous}, we incorporate instances labeled as \textit{not enough info} into our dataset. To this end, we either randomly eliminate evidence from the corresponding \textit{SUP/REF} instances or merge evidence from these two types to generate \textit{NEI} evidence.

$\bullet$ \textbf{Argument Structure Generation} \quad
To describe the logical reasoning process for each instance, we require GPT-4 to generate succinct argument structures for both the \textit{SUP} and \textit{REF} instances\footnote{Since the containing of non-gold evidence for \textit{NEI}, we abandon to generate the argument structure for \textit{NEI}.}. Inspired by \citet{wang2022code4struct}, we prompt GPT-4 with Python-style code examples that have strict format restrictions. These examples help establish a logical reasoning path for each instance, thus deepening the structural understanding of arguments.

In specific, by referring to the argument structure outlined in \cref{subsec: our argument structure}, we introduce the \texttt{FACT} class, which symbolizes an evidence node within the argument structure, and three distinct functions to regulate the generation of GPT-4.
(I) The \texttt{one\_to\_one\_linking} function (corresponding to \textit{series argument}) represents the deduction from one piece of evidence to another.
(II) The \texttt{one\_to\_multiple\_linking} function (corresponding to \textit{divergent argument}) illustrates the deduction from one piece of evidence to multiple pieces.
(III) The \texttt{multiple\_to\_one\_linking} function (corresponding to \textit{combined argument}) represents the reasoning from several pieces of evidence to support a single one.

$\bullet$ \textbf{Claim Generation} \quad
By taking the cause-effect pairs extracted from W\textsc{iki}W\textsc{hy} as input,
the GPT-4 is prompted with crafted examples to generate diverse causal claims linguistically without accessing external information.

Notably, to better simulate the real-world fact verification scenarios and boost the distinguishing capability of models to the misleading evidence,
we employ BM25 \cite{robertson2009probabilistic} to retrieve and extract 300 pieces of evidence from the entire dataset that show a higher degree of lexical overlap with the claim.
Then, we use BLEURT \cite{sellam2020bleurt} to select 8-12 pieces of evidence with close semantic similarity to serve as distractor evidence for \textit{SUP}, \textit{REF}, and \textit{NEI}.

\subsection{Manual Validation}
\label{subsec.manual_validation}
The uncontrollable generation of LLMs might result in unfaithful or invalid instances.
Thus,
we subsequently incorporate the human validation procedure for two purposes:
(I)
to revise and edit the content of generated instances by LLMs;
(II))
to critically review and filter out the low-quality instances.
The details are shown in Appendix \ref{Detailed Description of the Annotation Process}.

\paragraph{Manual Editing}
We ask a group of annotators with NLP backgrounds to edit and refine the generated instances.

$\bullet$ \textbf{Claim and Evidence} \quad
Annotators are required to review the claims and assess whether they accurately contain all details from the provided cause-effect pair and express clear causality. Then, annotators are instructed to refine the generated evidence by evaluating whether all evidence aligns with the label and is presented in concise language (e.g., without repetition or redundancy).

$\bullet$ \textbf{Argument Structure} \quad
The annotators are required to review and correct errors within argument structures by double-checking the validity of each reasoning step and removing unnecessary non-code content, such as comments. In this context, we encourage annotators to apply their judgment and expertise throughout the process.

\paragraph{Manual Filtering}
Another group of annotators is required to filter out the low-quality edited instances to further improve the quality of the dataset. Given the instructions and demonstration of the annotated argument structures, annotators need to decide whether to retain or discard instances based on the following three criteria.
Finally, 6,757 and 6,638 instances pass the filter for \textit{SUP} and \textit{REF}, respectively.

\noindent $\bullet$ \textbf{Faithfulness:} The truthfulness of the claim can be verified by the evidence and is consistent with the given label.

\noindent $\bullet$ \textbf{Fluency:} The claim and associated evidence are written fluently and without grammatical errors.

\noindent $\bullet$ \textbf{Validity:} The argument structure is logically coherent and correctly depicts the reasoning process from cause to effect.

\paragraph{Quality Control}
We apply strict quality control measures by following the principles outlined in the Dataset Statement \cite{bender2018data}, ensuring high-quality annotation. Specifically, for each instance, two annotators are required to perform the manual editing and filtering, and resolve any disagreements identified by a third annotator.
We employ majority voting to determine the final retained instances, reaching an inter-rater agreement at Fleiss’s $\kappa$ = 0.75~\citep{fleiss1981measurement}. The details are listed in Appendix \ref{Detailed Description of the Annotation Process}.

\section{Dataset Analysis}
\subsection{Dataset Statistics}
Table~\ref{table Statistics} presents the statistics of our \textsc{CheckWhy} dataset. This dataset comprises 6,532 sets of instances, with each set including a \textit{SUP} instance, a \textit{REF} instance, and an \textit{NEI} instance, resulting in 19,596 instances in total.
In addition, due to the lack of argument structure for \textit{NEI} --- the \textit{SUP} and \textit{REF} instances contain three elements \{\textit{claim}, \textit{evidence}, \textit{argument structure}\}, while the \textit{NEI} instance is limited to the \textit{claim} and \textit{evidence} ---
we build a \textsc{CheckWhy}$_2$ dataset by excluding \textit{NEI} instances to assess the reasoning ability through argument structures within \textit{SUP}/\textit{REF} instances.
Compared to \textsc{WikiWhy}, our \textsc{CheckWhy} includes more extended and detailed reasoning steps, with an average of 13.7 pieces of evidence and 3.14 reasoning steps. The histograms of evidence length and reasoning step counts are detailed in Appendix~\ref{ablation experiment}.

\begin{table}[t]
	\centering
	%\tiny
	\resizebox{0.95\linewidth}{!}{
		\renewcommand\arraystretch{1}
		\begin{tabular}{lcc}
			\toprule[1pt]
			%\multicolumn{3}{c}{\textbf{C\textsc{heck}W\textsc{hy} Statistics}}\\ \midrule[0.5pt]
			                            & \textsc{CheckWhy} & \textsc{CheckWhy}$_2$ \\ \midrule[0.5pt]
			\# of All                   & 19,596            & 13,062                \\ \midrule[0.5pt]
			\# of Train                 & 14,796            & 9,864                 \\
			\# of Dev                   & 2,400             & 1,600                 \\
			\# of Test                  & 2,400             & 1,600                 \\ \midrule[0.5pt]
			Avg. \# Evidence            & 11.1              & {13.7}                \\
			Avg. \# Distractor          & -                 & {9.5}                 \\
			Avg. \# Tokens per Claim    & 24.3              & {24.3}                \\
			Avg. \# Tokens per Evidence & 19.6              & {19.3}                \\
			Avg. \# Reasoning Steps     & -                 & {3.14}                \\
			\bottomrule[1pt]
		\end{tabular}}
	\caption{\label{table Statistics} Summary statistics of C\textsc{heck}W\textsc{hy}.
		Here,
		\textsc{CheckWhy} is the dataset containing \{\textit{SUP}, \textit{REF}, \textit{NEI}\} instances, while \textsc{CheckWhy}$_2$ is composed of \{\textit{SUP}, \textit{REF}\} instances.
		\textit{Avg. \#} denotes the number on average,
		and the \textit{Evidence} denotes the combination of the valid annotated evidence and the extracted distractor evidence.
	}
\end{table}

\subsection{Argument Structure Analysis}
Inspired by \citet{dalvi2021explaining},
to understand the intricacies of reasoning present within \textsc{CheckWhy},
we analyzed the reasoning steps extracted from 50 randomly selected instances with diverse argument structures.
We identified 5 prevalent high-level categories of reasoning, as shown in Table~\ref{5 common reasoning methods}.
\textit{Event Causality} (36.1\%) refers to the relations where one event directly leads to another event.
\textit{Inference from Properties} (22.2\%) involves concluding by referring to the properties present within one input.
\textit{Rule-based Inference} (17.3\%) is about deducing new information or making decisions grounded in a set of predefined rules or conditions.
\textit{Inductive Reasoning} (14.5\%) requires a model to make generalized conclusions based on specific observations or evidence, deriving general principles or patterns from particular examples or observations.
Moreover,
with 9.9\% of instances require applying \textit{Sequential Reasoning}.
Overall, this analysis reveals diverse forms of reasoning steps that are essential for inferring the argument structure in \textsc{CheckWhy}.
In addition, we summarize 5 types of prevalent complex argument structures, as shown in Figure~\ref{Types of argument diagrams},
which also shows the complexity of our argument structure through a macro perspective.

\begin{table*}[t]\small
	\small
	\centering
	\resizebox{\linewidth}{!}{
		\renewcommand\arraystretch{1}
		\begin{tabular}{llll}
			\toprule[1pt]
			\textbf{Inference Type}   & \textbf{Prop.} &       & \textbf{Example}                                                                                                        \\ \midrule[0.5pt]
			Event Causality           & 36.1\%         & $E_1$ & \textcolor{orange}{Heavy rain and flooding} have caused significant damage to the area.                                 \\
			                          &                & $E_2$ & \textcolor{blue}{Road closures and traffic disruptions} are affecting the entire city.                                  \\
			                          &                &       & $E_1$ $\rightarrow$ $E_2$                                                                                               \\ \midrule[0.5pt]
			Inference from Properties & 22.2\%         & $E_1$ & If the patient has \textcolor{orange}{a fever and cough}, diagnose them with \textcolor{blue}{a respiratory infection.} \\
			                          &                & $E_2$ & David has \textcolor{orange}{a fever and cough}.                                                                        \\
			                          &                & $E_3$ & he is likely suffering from \textcolor{blue}{a respiratory infection}.                                                  \\
			                          &                &       & $E_1$, $E_2$ $\rightarrow$ $E_3$                                                                                        \\ \midrule[0.5pt]
			Rule-based Inference      & 17.3\%         & $E_1$ & All \textcolor{red}{humans} are \textcolor{blue}{mortal}.                                                               \\
			                          &                & $E_2$ & \textcolor{orange}{Socrates} is a \textcolor{red}{human}.                                                               \\
			                          &                & $E_3$ & Therefore, \textcolor{orange}{Socrates} is \textcolor{blue}{mortal}.                                                    \\
			                          &                &       & $E_1$, $E_2$ $\rightarrow$ $E_3$                                                                                        \\ \midrule[0.5pt]
			Inductive Reasoning       & 14.5\%         & $E_1$ & \textcolor{red}{Every morning} for the past week, you've woken up to find the \textcolor{red}{grass wet}.               \\
			                          &                & $E_2$ & Based on these observations, you hypothesize that \textcolor{red}{it rains during the night}.                           \\
			                          &                &       & $E_1$ $\rightarrow$ $E_2$                                                                                               \\ \midrule[0.5pt]
			Sequential Reasoning      & 9.9\%          & $E_1$ & A heavy rainstorm has occurred in \textcolor{orange}{the capital of Italy}.                                             \\
			                          &                & $E_2$ & \textcolor{orange}{The capital of Italy} is \textcolor{blue}{Rome}.                                                     \\
			                          &                & $E_3$ & \textcolor{blue}{Rome} has experienced a heavy rainstorm.                                                               \\
			                          &                &       & $E_1$, $E_2$ $\rightarrow$ $E_3$                                                                                        \\

			\bottomrule[1pt]
		\end{tabular}
	}
	\caption{\label{Infer type} The prevalence of 5 reasoning steps required for reasoning on argument structure, sampled from 50 random instances in the training set. Here, $E_{n}$ denotes input evidence, and the ``$\rightarrow$'' symbolizes the ``reasoning''.}
	\label{5 common reasoning methods}
\end{table*}

\begin{table}[t]
	\small
	\centering
	\tiny
	\resizebox{\linewidth}{!}{
		\begin{tabular}{lcccccc}
			\toprule[1pt]
			\multicolumn{1}{l}{\multirow{2.5}{*}{\textbf{Model}}} & \multicolumn{2}{c}{\textbf{Task1 ($Y_3$)}} & \multicolumn{2}{c}{\textbf{Task1 ($Y_2$)}} & \multicolumn{2}{c}{\textbf{Task2}}                                                                         \\
			\cmidrule(lr){2-3}\cmidrule(lr){4-5}\cmidrule(lr){6-7}
			                                                      & Acc.                                       & F1                                         & Acc.                               & F1            & Acc.                      & F1                        \\
			\midrule[0.5pt]

			\textbf{\textit{Discriminative Models}}                                                                                                                                                                                                                      \\
			BERT (FT)                                             & \textbf{73.3}                              & \textbf{72.9}                              & 76.9                               & 76.9          & \underline{88.0}          & \underline{88.0}          \\
			Transformer-XH (FT)                                   & 63.2                                       & 62.2                                       & \textbf{78.9}                      & \textbf{78.9} & \underline{\textbf{90.4}} & \underline{\textbf{90.4}} \\
			UniXcoder (FT)                                        & 70.1                                       & 69.5                                       & 77.2                               & 77.2          & \underline{83.4}          & \underline{83.4}          \\
			\midrule[0.2pt]

			\textbf{\textit{Generative Models}}                                                                                                                                                                                                                          \\
			CodeT5 (FT)                                           & 72.3                                       & 72.0                                       & 71.6                               & 71.6          & \underline{78.0}          & \underline{78.0}          \\
			CodeT5+ (FT)                                          & 72.8                                       & 72.5                                       & 65.6                               & 65.4          & \underline{79.3}          & \underline{79.3}          \\
			\midrule[0.2pt]

			\textbf{\textit{Large Language Models}}                                                                                                                                                                                                                      \\
			ChatGPT (CoT)                                         & 37.8                                       & 31.1                                       & 60.6                               & 60.7          & \underline{64.4}          & 60.7                      \\
			GPT-4 (CoT)                                           & 49.6                                       & 41.0                                       & 62.5                               & 62.3          & \underline{72.8}          & \underline{71.0}
			\\ \bottomrule[1pt]
		\end{tabular}}
	\caption{ The performance of baselines on Task 1 and Task 2, where FT denotes that the model is fine-tuned on \textsc{CheckWhy} training set. The CoT denotes the Chain-of-Thought prompts with few examples. The best results are marked in \textbf{bold} and the results with further improvement after the incorporation of {argument structures} are \underline{underlined}.}
	\label{table result task12}
\end{table}

\section{Experiment}

\paragraph{Task Notation}

The instance within \textsc{CheckWhy} is denoted by the quadruple $(C, E/LE, S, Y_2 | Y_3)$, where $C$ denotes the claim, $E=\{e_1, e_2, \dots, e_n\}$ denotes the evidence, $LE$ refers to the initial leaf evidence in an argument structure. Notably, we ensure that $LE$ includes the valid leaf evidence and all the distractor evidence, which requires the denoising capability of models. In addition, $S$ denotes the argument structure, and $Y_3 \in \{REF, SUP, NEI\}$ or $Y_2 \in \{REF, SUP\}$ for \textsc{CheckWhy}$_2$.
Based on our \textsc{CheckWhy}, we define four tasks of increasing difficulty, with the aim of (i) predicting whether the evidence supports or refutes the claim, or presents not enough information; (ii) generating valid argument structures outlining the reasoning process for causal fact verification. The following describes four tasks.

\paragraph{Task 1: Input = $(C, LE)$, Output = $Y_2|Y_3.$}
We follow the traditional fact verification formulation to evaluate the inference capability of models on the C\textsc{heck}W\textsc{hy} dataset, i.e.,
verifying the claim based on foundational leaf evidence.
In this vein, we conduct the experiments on three types of baselines: (I) Discriminative models: BERT \cite{devlin2018bert}, Transformer-XH \cite{zhao2019transformer}, and UniXcoder \cite{guo2022unixcoder}, (II) Generative models: CodeT5 \cite{wang2021codet5} and CodeT5+ \cite{wang2023codet5+}, (III) LLMs: ChatGPT and GPT-4 \footnote{https://platform.openai.com/docs/models/gpt-4-and-gpt-4-turbo}\cite{achiam2023gpt} with Chain-of-Thought prompts \citep{wei2022chain}.

\paragraph{Task 2: Input = $(C, LE, S)$, Output = $Y_2.$}
Task 2 is designed to assess whether models have a better understanding when incorporating the structural information within the argument structure during the verification.
In this vein, We adopt the same set of baselines as in Task 1.

\paragraph{Task 3: Input = $(C, E)$, Output = $(Y_2,S).$}
To investigate the logical reasoning ability of existing models when performing verification, we experiment that involves the simultaneous verification of the claim and the generation of the argument structure.
Here, we conduct experiments on CodeT5, CodeT5+, GPT-4 and ChatGPT with Chain-of-Thought prompts.

\paragraph{Task 4: Input = $(C, LE)$, Output = $(Y_2,S).$}
Task 4 investigates whether the model can independently construct an entire argument structure from bottom to top based on leaf evidence and its reasoning ability,
which is the most challenging experimental setup in this paper. The model must generate the verification label and a complete argument structure based on the input claim and leaf evidence nodes. We use the same baseline models as in Task 3 for this task.

\subsection{Evaluation Metrics}

\paragraph{Automatic Evaluation Metircs}
Due to the uniqueness of argument structure within \textsc{CheckWhy},
traditional evaluation metrics in text generation may not be suitable in our setting.
Thus, we propose two new evaluation metrics focusing on argument structure generation.

$\bullet$ \textbf{Structure Similarity} \quad Inspired by \citet{saha2021explagraphs}, we introduce an automated evaluation metric named Structure Similarity. The metric treats the argument structure as a collection of edges, with each edge considered as a sentence, and a matching method is employed to determine the optimal alignment between the edges in the predicted structure and those in the gold argument structure.
In our experiments, the BERTScore based on DeBERTa-large \cite{he2020deberta} is used as the scoring function to measure how closely the predicted edges match the gold edges. The detailed algorithm can be found in \citet{saha2021explagraphs}.

$\bullet$ \textbf{Exact Match Similarity} \quad Similar to structure similarity, exact match similarity employs a stricter matching strategy.
Here, structures are also regarded as sequences of edges, but a match is considered successful only when the predicted edge exactly matches the gold edge.

\paragraph{Human Evaluation Criteria}
To improve the reliability of the evaluation on argument structure, we introduce extra human evaluations beyond the automated evaluation metircs.
Empirically, we randomly select 50 instances and ask three graduate students with NLP background to assess the generated argument structure according to the following criteria with binary score.

\noindent $\bullet$ \textbf{Validity:} Is the generated argument structure logically coherent?

\noindent $\bullet$ \textbf{Win/Tie/Lose:} Comparing the generated argument structure against the provided reference. Mark \textit{Win} if you prefer the generated structure, \textit{Tie} if you have no preference, and \textit{Lose} if you prefer the reference structure.

\begin{table*}[t]
	\small
	\centering
	\resizebox{0.95\linewidth}{!}{
		\begin{tabular}{cccccccccc}
			\toprule[1pt]
			% \multicolumn{2}{l}{\multirow{2.5}{*}{\textbf{Model}}} & \multicolumn{2}{c}{\textbf{Automatic}} & \multicolumn{4}{c}{\textbf{Human}} & \multicolumn{2}{c}{\textbf{Verification}} \\
			\multicolumn{2}{c}{\multirow{2.5}{*}{\textbf{Model}}} & \multicolumn{2}{c}{\textbf{Automatic}}                      & \multicolumn{4}{c}{\textbf{Human}} & \multicolumn{2}{c}{\textbf{Verification}}                                                                                                               \\
			\cmidrule(lr){3-4}\cmidrule(lr){5-8}\cmidrule(lr){9-10}
			\multicolumn{2}{c}{}
			                                                      & SS ($\uparrow$)                                             & ES ($\uparrow$)                    & Win ($\uparrow$)                          & Tie           & Lose ($\downarrow$) & Validity ($\uparrow$) & Acc.          & F1                            \\
			\midrule[0.7pt]
			\multirow{7}{*}{\textbf{{Task 3}}}                    & \multicolumn{9}{l}{\textit{\textbf{Generative Models}}}                                                                                                                                                                                                    \\
			                                                      & CodeT5 (FT)                                                 & 87.4                               & 48.4                                      & 10.0          & 27.5                & 62.5                  & 52.5          & 83.7          & 83.5          \\
			                                                      & CodeT5+ (FT)                                                & \textbf{88.8}                      & \textbf{53.4}                             & 7.5           & 40.0                & 52.5                  & \textbf{62.5} & \textbf{86.6} & \textbf{86.6} \\
			\cmidrule[0.2pt](r){2-10}
			                                                      & \multicolumn{9}{l}{\textit{\textbf{Large Language Models}}}                                                                                                                                                                                                \\
			                                                      & ChatGPT (CoT)                                               & 69.5                               & 26.4                                      & 0.0           & 17.5                & 82.5                  & 37.5          & 63.4          & 58.5          \\
			                                                      & GPT-4 (CoT)                                                 & 77.7                               & 34.2                                      & \textbf{21.6} & \textbf{40.5}       & \textbf{37.8}         & 62.2          & 72.5          & 70.9          \\

			\midrule[0.7pt]
			\multirow{7}{*}{\textbf{{Task 4}}}                    & \multicolumn{9}{l}{\textbf{\textit{Generative Models}}}                                                                                                                                                                                                    \\
			                                                      & CodeT5 (FT)                                                 & 72.8                               & 1.1                                       & 5.5           & 30.3                & 64.2                  & 55.2          & 68.9          & 67.9          \\
			                                                      & CodeT5+ (FT)                                                & \textbf{75.7}                      & \textbf{2.5}                              & 6.1           & \textbf{33.9}       & \textbf{60.0}         & \textbf{59.5} & \textbf{72.6} & \textbf{72.6} \\
			\cmidrule[0.2pt](r){2-10}
			                                                      & \multicolumn{9}{l}{\textbf{\textit{Large Language Models}}}                                                                                                                                                                                                \\
			                                                      & ChatGPT (CoT)                                               & 60.9                               & 1.8                                       & 4.2           & 28.5                & 67.3                  & 52.9          & 53.9          & 44.7          \\
			                                                      & GPT-4 (CoT)                                                 & 66.7                               & 1.4                                       & \textbf{6.2}  & 33.2                & 60.6                  & 57.8          & 57.0          & 47.8          \\

			\bottomrule[1pt]
		\end{tabular}}
	\caption{\label{table result task34} The performance of baselines on Task 3 and Task 4, where
		SS denotes structure similarity and ES denotes exact match similarity. The best results are marked in \textbf{bold}.}
\end{table*}

\subsection{Results and Discussion}

\paragraph{Main Results}

The overall results of baselines on our CHECKWHY dataset are reported in Table~\ref{table result task12} and Table~\ref{table result task34}.
In general, for Task 1 and Task 2, generative models show notably poorer performance compared to the discriminative models,
and LLMs encounter substantial challenges in causal verification.
Despite the powerful capability of GPT-4, which shows an improvement over ChatGPT for verification, it still lags behind the top-performing fine-tuned models.
In addition, through the comparison of $Y_3$ and $Y_2$ on Task 1,
we observe the consistent improvement for both discriminative models and LLMs, which indicates the disruptive impact brought by the ambiguous evidence within \textit{NEI} instance, particularly on LLMs.
An interesting observation here is that there is a slight decrease in the performance of generative models, which may be attributed to the disparities between the datasets utilized during the pretraining and fine-tuning phases.
For Task 3 and Task 4, compared to the generative models,
a similar phenomenon observed is that LLMs struggle to achieve satisfactory results in the generation of argument structure and show notably poorer performance on causal verification. This may be due to the conflict raised by the internal world knowledge and the evidence we generated, e.g., there is evidence that contradicts the real-world situation.
In addition, the difference can be observed in the human evaluation, where LLMs present a slightly higher performance compared to the generative models.
This may be attributed to the fact that LLMs generate more fluency and human-readable text with the powerful generative capability.
Overall, the results of four tasks validate the difficulty when facing the setting of our \textsc{CheckWhy}.

\paragraph{The Effectiveness of Argument Structures}

Argument structures prove to be particularly advantageous, especially in specific experimental setups. In Task 1, where structures are absent, models such as the CodeT5 series and LLMs, relying solely on textual information, may struggle to achieve high accuracy.
Furthermore, when comparing Task 1 and Task 2, a consistent improvement is noted across all baselines if structures are directly provided. For example, GPT-4 achieves approximately a 10\% increase in accuracy in Task 2 compared to Task 1. This effect shows the most prominent performance on Transformer-XH, a GNN method conducting reasoning over the evidence graph.
This underscores the significance of structural information within the argument structure for causal fact verification.

\paragraph{The Construction of Argument Structures}
In Task 3 and Task 4, we investigate whether models can generate argument structures. Structure similarity may be significantly higher than exact match similarity because structure similarity considers the semantic meaning within nodes, whereas exact match similarity only checks whether characters are identical.
CodeT5+ exhibits particularly noticeable performance, achieving the highest exact match similarity and structure similarity, as well as the highest F1 score and accuracy.
Task 4 investigates whether models can construct an entire structure and complete inference given only the leaf nodes. All models perform poorly on Task 4, which may be attributed to the vast generation space in creating an entire structure.
This indicates that when all evidence nodes are provided, the model only needs to connect these nodes, instead of inferring the complete reasoning path. However, when only leaf nodes are provided, the task becomes much more challenging.

\paragraph{Human Evaluation}
Our human evaluation experiments, as detailed in Table~\ref{table result task34}, reveal that there is significant room for improvement across different aspects. None of the baseline models, including CodeT5, CodeT5+, and GPT-4, exhibit notably superior performance. Specifically, our strongest baseline, CodeT5+, generates valid argument structures only 62.5\% of the time in Task 3, whereas in Task 4 it generates up to 60.0\% of argument structures that are worse than the gold references. These results from the strongest LLMs leave ample room for improvement and serve as motivation for future work on these tasks.

\section{Related work}
\subsection{Fact Verification Datasets}
The fact verification task has recently garnered significant attention within the NLP community.
Researchers prompt the development of fact verification by introducing various resources.
Well-known datasets such as FEVER \citep{thorne2018fever}, PolitiFact \citep{garg2020new}, and CREAK \citep{onoe2021creak} regard the fact verification task as a form of natural language inference, aiming to predict whether the evidence supports or contradicts a claim.
Subsequently, some datasets have been proposed to address fact-checking for complex claims necessitating multi-step reasoning \cite{aly2021feverous, jiang2020hover}, which typically present multiple pieces of evidence linked to the claim through entity matching.
Additionally, to improve the interpretability of the veracity predictions, \citet{alhindi2018your} extends the LIAR dataset by incorporating summaries from fact-checking articles. Furthermore, \citet{kotonya2020explainable} introduces the first dataset explicitly containing gold explanations, composed of fact-checking articles and other news items. \citet{rani2023factify} leverages interrogative questions to enhance interpretability by exploring the semantic factoids within the claim.

Despite the advancement, existing datasets are primarily designed for verifying the semantic factoids with limited types within claims.
Our work for the first time focuses on ``why'' claims: verifying the causal relation within the claim rather than basic semantic factoids. In addition, inspired by the literature on logical theory, we construct argument structures to explicitly represent the reasoning process, enhancing the causal reasoning ability and interpretability of the model.

\subsection{Causal Reasoning}
With the increasing attention on causality \cite{willig2022probing, zhang2023understanding}, several formulations of causality-related skills for NLP are proposed, which can be summarized into (1) causality as knowledge \cite{sap2019atomic}, encompasses the representation, understanding, and utilization of causal relationships embedded within textual data, (2) causality as language comprehension \cite{stede2008connective, cao2022cognitive,  yu2019detecting}, originates from traditional linguistic studies on causal connectives and the usage of causal language, extending to more recent efforts in causal relation extraction, and (3) causality as reasoning \citep{NEURIPS2023_631bb943}, involves identifying factors leading to outcomes and understanding underlying mechanisms. Our work is the first attempt to introduce causality into the fact verification task inspired by the causal reasoning in argumentation \cite{habernal-etal-2018-argument, heinisch-etal-2022-overview}, in which causal knowledge, causal language comprehension, and causal reasoning are significant. Exploring whether models can grasp the causal relationship between evidence and claim in the verification process will further promote the development of the task.

\section{Conclusion}
In this paper, we introduce \textsc{CheckWhy}, a challenging dataset and benchmark built around a novel causal fact verification task, annotated by a human-model collaborative approach.
By drawing inspiration from the logical theory, we incorporate the argument structure to represent the explicit logical reasoning process when assessing the causal relation within the ``why'' claim, which may prove valuable for developing multi-step reasoning skills across various scenarios.
Extensive experiments on various state-of-the-art models validate the importance of incorporating argument structures and the difficulty of generating them. With the baselines achieving limited results, we believe that \textsc{CheckWhy} is a challenging yet attractive benchmark for the development of fact verification systems.

\section*{Limitations}
$\bullet$ Within the \textsc{CheckWhy} framework, the label assigned to each claim is based on distinct argument structures. Therefore, the assigned label may not always correspond to real-world circumstances.

\noindent $\bullet$ The \textsc{CheckWhy} dataset is created by LLMs, which could face difficulties in retrieving evidence in open-domain scenarios compared to previous datasets. This includes selecting relevant evidence in a broad and unrestricted information space.

\noindent $\bullet$ The initial evidence in the argument structure, as provided by LLMs, may not meet everyone's expectations or align with their understanding. This could affect the universal acceptability and perceived validity of the evidence.

\section*{Acknowledgement}

The authors would like to thank the anonymous
reviewers for their insightful comments. This work
is funded by the National Natural Science Foundation of China (Grant No.62176053, No.62376130), Shandong Provincial Natural Science Foundation (Grant No.ZR2022MF243), Program of New Twenty Policies for Universities of Jinan (Grant No.202333008), and supported by the Big Data Computing Center of Southeast
University.

\normalem
\bibliography{custom}

\begin{thebibliography}{57}
\expandafter\ifx\csname natexlab\endcsname\relax\def\natexlab#1{#1}\fi

\bibitem[{Achiam et~al.(2023)Achiam, Adler, Agarwal, Ahmad, Akkaya, Aleman, Almeida, Altenschmidt, Altman, Anadkat et~al.}]{achiam2023gpt}
Josh Achiam, Steven Adler, Sandhini Agarwal, Lama Ahmad, Ilge Akkaya, Florencia~Leoni Aleman, Diogo Almeida, Janko Altenschmidt, Sam Altman, Shyamal Anadkat, et~al. 2023.
\newblock Gpt-4 technical report.
\newblock \emph{arXiv preprint arXiv:2303.08774}.

\bibitem[{Alhindi et~al.(2018)Alhindi, Petridis, and Muresan}]{alhindi2018your}
Tariq Alhindi, Savvas Petridis, and Smaranda Muresan. 2018.
\newblock \href {https://doi.org/10.18653/v1/W18-5513} {Where is your evidence: Improving fact-checking by justification modeling}.
\newblock In \emph{Proceedings of the First Workshop on Fact Extraction and {VER}ification ({FEVER})}, pages 85--90.

\bibitem[{Aly et~al.(2021)Aly, Guo, Schlichtkrull, Thorne, Vlachos, Christodoulopoulos, Cocarascu, and Mittal}]{aly2021feverous}
Rami Aly, Zhijiang Guo, Michael~Sejr Schlichtkrull, James Thorne, Andreas Vlachos, Christos Christodoulopoulos, Oana Cocarascu, and Arpit Mittal. 2021.
\newblock \href {https://doi.org/10.18653/v1/2021.fever-1.1} {The fact extraction and {VER}ification over unstructured and structured information ({FEVEROUS}) shared task}.
\newblock In \emph{Proceedings of the Fourth Workshop on Fact Extraction and VERification (FEVER)}.

\bibitem[{Bender and Friedman(2018)}]{bender2018data}
Emily~M. Bender and Batya Friedman. 2018.
\newblock \href {https://doi.org/10.1162/tacl_a_00041} {Data statements for natural language processing: Toward mitigating system bias and enabling better science}.
\newblock \emph{Transactions of the Association for Computational Linguistics}, 6:587--604.

\bibitem[{Cao et~al.(2022)Cao, Williamson, and Choi}]{cao2022cognitive}
Angela Cao, Gregor Williamson, and Jinho~D. Choi. 2022.
\newblock \href {https://aclanthology.org/2022.law-1.18} {A cognitive approach to annotating causal constructions in a cross-genre corpus}.
\newblock In \emph{Proceedings of the Linguistic Annotation Workshop (LAW-XVI)}, pages 151--159.

\bibitem[{Chen et~al.(2023)Chen, Gao, Bosselut, Sabharwal, and Richardson}]{chen-etal-2023-disco}
Zeming Chen, Qiyue Gao, Antoine Bosselut, Ashish Sabharwal, and Kyle Richardson. 2023.
\newblock \href {https://doi.org/10.18653/v1/2023.acl-long.302} {{DISCO}: Distilling counterfactuals with large language models}.
\newblock In \emph{Proceedings of the Annual Meeting of the Association for Computational Linguistics}, pages 5514--5528.

\bibitem[{Dalvi et~al.(2021)Dalvi, Jansen, Tafjord, Xie, Smith, Pipatanangkura, and Clark}]{dalvi2021explaining}
Bhavana Dalvi, Peter Jansen, Oyvind Tafjord, Zhengnan Xie, Hannah Smith, Leighanna Pipatanangkura, and Peter Clark. 2021.
\newblock \href {https://doi.org/10.18653/v1/2021.emnlp-main.585} {Explaining answers with entailment trees}.
\newblock In \emph{Proceedings of the Conference on Empirical Methods in Natural Language Processing}, pages 7358--7370.

\bibitem[{Devlin et~al.(2019)Devlin, Chang, Lee, and Toutanova}]{devlin2018bert}
Jacob Devlin, Ming-Wei Chang, Kenton Lee, and Kristina Toutanova. 2019.
\newblock \href {https://doi.org/10.18653/v1/N19-1423} {{BERT}: Pre-training of deep bidirectional transformers for language understanding}.
\newblock In \emph{Proceedings of the Conference of the North {A}merican Chapter of the Association for Computational Linguistics}, pages 4171--4186.

\bibitem[{Fleiss et~al.(1981)Fleiss, Levin, Paik et~al.}]{fleiss1981measurement}
Joseph~L Fleiss, Bruce Levin, Myunghee~Cho Paik, et~al. 1981.
\newblock The measurement of interrater agreement.
\newblock \emph{Statistical methods for rates and proportions}, 2(212-236):22--23.

\bibitem[{Freeman(2011)}]{freeman2011argument}
James~B Freeman. 2011.
\newblock \emph{Argument Structure:: Representation and Theory}.
\newblock Springer Science \& Business Media.

\bibitem[{Garg and Sharma(2020)}]{garg2020new}
Sonal Garg and Dilip~Kumar Sharma. 2020.
\newblock \href {https://doi.org/10.1109/SMART50582.2020.9337152} {New politifact: A dataset for counterfeit news}.
\newblock In \emph{International Conference System Modeling and Advancement in Research Trends (SMART)}, pages 17--22.

\bibitem[{Glockner et~al.(2022)Glockner, Hou, and Gurevych}]{DBLP:conf/emnlp/Glockner0G22}
Max Glockner, Yufang Hou, and Iryna Gurevych. 2022.
\newblock \href {https://doi.org/10.18653/V1/2022.EMNLP-MAIN.397} {Missing counter-evidence renders {NLP} fact-checking unrealistic for misinformation}.
\newblock In \emph{Proceedings of the Conference on Empirical Methods in Natural Language Processing}, pages 5916--5936.

\bibitem[{Glockner et~al.(2021)Glockner, Staliunaite, Thorne, Vallejo, Vlachos, and Gurevych}]{Glockner2021AmbiFCFA}
Max Glockner, Ieva Staliunaite, James Thorne, Gisela Vallejo, Andreas Vlachos, and Iryna Gurevych. 2021.
\newblock \href {https://api.semanticscholar.org/CorpusID:258987607} {Ambifc: Fact-checking ambiguous claims with evidence}.
\newblock \emph{Transactions of the Association for Computational Linguistics}, 12:1--18.

\bibitem[{Grimshaw(1990)}]{grimshaw1990argument}
Jane Grimshaw. 1990.
\newblock \emph{Argument structure.}
\newblock the MIT Press.

\bibitem[{Guo et~al.(2022{\natexlab{a}})Guo, Lu, Duan, Wang, Zhou, and Yin}]{guo2022unixcoder}
Daya Guo, Shuai Lu, Nan Duan, Yanlin Wang, Ming Zhou, and Jian Yin. 2022{\natexlab{a}}.
\newblock \href {https://doi.org/10.18653/v1/2022.acl-long.499} {{U}ni{X}coder: Unified cross-modal pre-training for code representation}.
\newblock In \emph{Proceedings of the Annual Meeting of the Association for Computational Linguistics}, pages 7212--7225.

\bibitem[{Guo et~al.(2022{\natexlab{b}})Guo, Schlichtkrull, and Vlachos}]{DBLP:journals/tacl/GuoSV22}
Zhijiang Guo, Michael~Sejr Schlichtkrull, and Andreas Vlachos. 2022{\natexlab{b}}.
\newblock \href {https://doi.org/10.1162/TACL\_A\_00454} {A survey on automated fact-checking}.
\newblock \emph{Trans. Assoc. Comput. Linguistics}, pages 178--206.

\bibitem[{Habernal et~al.(2018)Habernal, Wachsmuth, Gurevych, and Stein}]{habernal-etal-2018-argument}
Ivan Habernal, Henning Wachsmuth, Iryna Gurevych, and Benno Stein. 2018.
\newblock \href {https://doi.org/10.18653/v1/N18-1175} {The argument reasoning comprehension task: Identification and reconstruction of implicit warrants}.
\newblock In \emph{Proceedings of the Conference of the North {A}merican Chapter of the Association for Computational Linguistics: Human Language Technologies}, pages 1930--1940.

\bibitem[{He et~al.(2021)He, Liu, Gao, and Chen}]{he2020deberta}
Pengcheng He, Xiaodong Liu, Jianfeng Gao, and Weizhu Chen. 2021.
\newblock \href {https://openreview.net/forum?id=XPZIaotutsD} {Deberta: Decoding-enhanced bert with disentangled attention}.
\newblock In \emph{Proceedings of the International Conference on Learning Representations}.

\bibitem[{Heinisch et~al.(2022)Heinisch, Frank, Opitz, Plenz, and Cimiano}]{heinisch-etal-2022-overview}
Philipp Heinisch, Anette Frank, Juri Opitz, Moritz Plenz, and Philipp Cimiano. 2022.
\newblock \href {https://aclanthology.org/2022.argmining-1.7} {Overview of the 2022 validity and novelty prediction shared task}.
\newblock In \emph{Proceedings of the Workshop on Argument Mining}, pages 84--94.

\bibitem[{Ho et~al.(2023)Ho, Sharma, Chang, Saxon, Levy, Lu, and Wang}]{ho2022wikiwhy}
Matthew Ho, Aditya Sharma, Justin Chang, Michael Saxon, Sharon Levy, Yujie Lu, and William~Yang Wang. 2023.
\newblock \href {https://openreview.net/forum?id=vaxnu-Utr4l} {Wikiwhy: Answering and explaining cause-and-effect questions}.
\newblock In \emph{Proceedings of the International Conference on Learning Representations}.

\bibitem[{Jiang et~al.(2020)Jiang, Bordia, Zhong, Dognin, Singh, and Bansal}]{jiang2020hover}
Yichen Jiang, Shikha Bordia, Zheng Zhong, Charles Dognin, Maneesh Singh, and Mohit Bansal. 2020.
\newblock \href {https://doi.org/10.18653/v1/2020.findings-emnlp.309} {{H}o{V}er: A dataset for many-hop fact extraction and claim verification}.
\newblock In \emph{Findings of the Association for Computational Linguistics: EMNLP}, pages 3441--3460.

\bibitem[{Jin et~al.(2023)Jin, Chen, Leeb, Gresele, Kamal, LYU, Blin, Gonzalez~Adauto, Kleiman-Weiner, Sachan, and Sch\"{o}lkopf}]{NEURIPS2023_631bb943}
Zhijing Jin, Yuen Chen, Felix Leeb, Luigi Gresele, Ojasv Kamal, Zhiheng LYU, Kevin Blin, Fernando Gonzalez~Adauto, Max Kleiman-Weiner, Mrinmaya Sachan, and Bernhard Sch\"{o}lkopf. 2023.
\newblock \href {https://proceedings.neurips.cc/paper_files/paper/2023/file/631bb9434d718ea309af82566347d607-Paper-Conference.pdf} {Cladder: Assessing causal reasoning in language models}.
\newblock In \emph{Advances in Neural Information Processing Systems}.

\bibitem[{Jin et~al.(2024)Jin, Liu, Lyu, Poff, Sachan, Mihalcea, Diab, and Sch{\"o}lkopf}]{jin2023can}
Zhijing Jin, Jiarui Liu, Zhiheng Lyu, Spencer Poff, Mrinmaya Sachan, Rada Mihalcea, Mona Diab, and Bernhard Sch{\"o}lkopf. 2024.
\newblock Can large language models infer causation from correlation?
\newblock In \emph{Proceedings of the International Conference on Learning Representations}.

\bibitem[{Kotonya and Toni(2020)}]{kotonya2020explainable}
Neema Kotonya and Francesca Toni. 2020.
\newblock \href {https://doi.org/10.18653/v1/2020.emnlp-main.623} {Explainable automated fact-checking for public health claims}.
\newblock In \emph{Proceedings of the Conference on Empirical Methods in Natural Language Processing and International Joint Conference on Natural Language Processing}, pages 7740--7754.

\bibitem[{Krishna et~al.(2022)Krishna, Riedel, and Vlachos}]{DBLP:journals/tacl/Krishna0022}
Amrith Krishna, Sebastian Riedel, and Andreas Vlachos. 2022.
\newblock \href {https://transacl.org/ojs/index.php/tacl/article/view/3527} {Proofver: Natural logic theorem proving for fact verification}.
\newblock \emph{Transactions of the Association for Computational Linguistics}, 10:1013--1030.

\bibitem[{Lewandowsky et~al.(2020)Lewandowsky, Cook, Ecker, Albarrac{\'\i}n, Amazeen, Kendeou, Lombardi, Newman, Pennycook, Porter et~al.}]{lewandowsky2020debunking}
Stephan Lewandowsky, John Cook, UKH Ecker, D~Albarrac{\'\i}n, MA~Amazeen, P~Kendeou, D~Lombardi, EJ~Newman, G~Pennycook, E~Porter, et~al. 2020.
\newblock The debunking handbook.
\newblock \emph{The Debunking Handbook}, pages 13--29.

\bibitem[{Lu et~al.(2023)Lu, Pan, Liu, Nakov, and Kan}]{DBLP:conf/emnlp/LuPLNK23}
Xinyuan Lu, Liangming Pan, Qian Liu, Preslav Nakov, and Min{-}Yen Kan. 2023.
\newblock \href {https://aclanthology.org/2023.emnlp-main.483} {{SCITAB:} {A} challenging benchmark for compositional reasoning and claim verification on scientific tables}.
\newblock In \emph{Proceedings of the Conference on Empirical Methods in Natural Language Processing}, pages 7787--7813.

\bibitem[{Onoe et~al.(2021)Onoe, Zhang, Choi, and Durrett}]{onoe2021creak}
Yasumasa Onoe, Michael Zhang, Eunsol Choi, and Greg Durrett. 2021.
\newblock \href {https://datasets-benchmarks-proceedings.neurips.cc/paper_files/paper/2021/file/5737c6ec2e0716f3d8a7a5c4e0de0d9a-Paper-round2.pdf} {Creak: A dataset for commonsense reasoning over entity knowledge}.
\newblock In \emph{Proceedings of the Neural Information Processing Systems Track on Datasets and Benchmarks}.

\bibitem[{Ouyang et~al.(2022)Ouyang, Wu, Jiang, Almeida, Wainwright, Mishkin, Zhang, Agarwal, Slama, Ray, Schulman, Hilton, Kelton, Miller, Simens, Askell, Welinder, Christiano, Leike, and Lowe}]{ouyang2022training}
Long Ouyang, Jeffrey Wu, Xu~Jiang, Diogo Almeida, Carroll Wainwright, Pamela Mishkin, Chong Zhang, Sandhini Agarwal, Katarina Slama, Alex Ray, John Schulman, Jacob Hilton, Fraser Kelton, Luke Miller, Maddie Simens, Amanda Askell, Peter Welinder, Paul~F Christiano, Jan Leike, and Ryan Lowe. 2022.
\newblock \href {https://proceedings.neurips.cc/paper_files/paper/2022/file/b1efde53be364a73914f58805a001731-Paper-Conference.pdf} {Training language models to follow instructions with human feedback}.
\newblock In \emph{Advances in Neural Information Processing Systems}, volume~35, pages 27730--27744.

\bibitem[{Pan et~al.(2023)Pan, Lu, Kan, and Nakov}]{DBLP:conf/emnlp/PanLKN23}
Liangming Pan, Xinyuan Lu, Min{-}Yen Kan, and Preslav Nakov. 2023.
\newblock \href {https://aclanthology.org/2023.emnlp-demo.23} {Qacheck: {A} demonstration system for question-guided multi-hop fact-checking}.
\newblock In \emph{Proceedings of the Conference on Empirical Methods in Natural Language Processing}, pages 264--273.

\bibitem[{Qi et~al.(2020)Qi, Zhang, Zhang, Bolton, and Manning}]{qi2020stanza}
Peng Qi, Yuhao Zhang, Yuhui Zhang, Jason Bolton, and Christopher~D. Manning. 2020.
\newblock \href {https://doi.org/10.18653/v1/2020.acl-demos.14} {{S}tanza: A python natural language processing toolkit for many human languages}.
\newblock In \emph{Proceedings of the Annual Meeting of the Association for Computational Linguistics}, pages 101--108.

\bibitem[{Rani et~al.(2023)Rani, Tonmoy, Dalal, Gautam, Chakraborty, Chadha, Sheth, and Das}]{rani2023factify}
Anku Rani, S.M Towhidul~Islam Tonmoy, Dwip Dalal, Shreya Gautam, Megha Chakraborty, Aman Chadha, Amit Sheth, and Amitava Das. 2023.
\newblock \href {https://doi.org/10.18653/v1/2023.acl-long.581} {{FACTIFY}-5{WQA}: 5{W} aspect-based fact verification through question answering}.
\newblock In \emph{Proceedings of the 61st Annual Meeting of the Association for Computational Linguistics}, pages 10421--10440.

\bibitem[{Robertson and Zaragoza(2009)}]{robertson2009probabilistic}
Stephen Robertson and Hugo Zaragoza. 2009.
\newblock \href {https://doi.org/10.1561/1500000019} {The probabilistic relevance framework: Bm25 and beyond}.
\newblock \emph{Foundations and Trends® in Information Retrieval}, 3(4):333--389.

\bibitem[{Romanou et~al.(2023)Romanou, Montariol, Paul, Laugier, Aberer, and Bosselut}]{romanou2023crab}
Angelika Romanou, Syrielle Montariol, Debjit Paul, Leo Laugier, Karl Aberer, and Antoine Bosselut. 2023.
\newblock \href {https://doi.org/10.18653/v1/2023.emnlp-main.940} {{CRAB}: Assessing the strength of causal relationships between real-world events}.
\newblock In \emph{Proceedings of the Conference on Empirical Methods in Natural Language Processing}, pages 15198--15216.

\bibitem[{Saha et~al.(2021)Saha, Yadav, Bauer, and Bansal}]{saha2021explagraphs}
Swarnadeep Saha, Prateek Yadav, Lisa Bauer, and Mohit Bansal. 2021.
\newblock \href {https://doi.org/10.18653/v1/2021.emnlp-main.609} {{E}xpla{G}raphs: An explanation graph generation task for structured commonsense reasoning}.
\newblock In \emph{Proceedings of the Conference on Empirical Methods in Natural Language Processing}, pages 7716--7740.

\bibitem[{Sap et~al.(2019)Sap, Le~Bras, Allaway, Bhagavatula, Lourie, Rashkin, Roof, Smith, and Choi}]{sap2019atomic}
Maarten Sap, Ronan Le~Bras, Emily Allaway, Chandra Bhagavatula, Nicholas Lourie, Hannah Rashkin, Brendan Roof, Noah~A. Smith, and Yejin Choi. 2019.
\newblock \href {https://doi.org/10.1609/aaai.v33i01.33013027} {Atomic: an atlas of machine commonsense for if-then reasoning}.
\newblock In \emph{Proceedings of the AAAI Conference on Artificial Intelligence and Innovative Applications of Artificial Intelligence Conference}, pages 3027--3035.

\bibitem[{Schuster et~al.(2019)Schuster, Shah, Yeo, Filizzola, Santus, and Barzilay}]{DBLP:conf/emnlp/SchusterSYFSB19}
Tal Schuster, Darsh~J. Shah, Yun Jie~Serene Yeo, Daniel Filizzola, Enrico Santus, and Regina Barzilay. 2019.
\newblock \href {https://doi.org/10.18653/V1/D19-1341} {Towards debiasing fact verification models}.
\newblock In \emph{Proceedings of the Conference on Empirical Methods in Natural Language Processing}, pages 3417--3423.

\bibitem[{Sellam et~al.(2020)Sellam, Das, and Parikh}]{sellam2020bleurt}
Thibault Sellam, Dipanjan Das, and Ankur Parikh. 2020.
\newblock \href {https://doi.org/10.18653/v1/2020.acl-main.704} {{BLEURT}: Learning robust metrics for text generation}.
\newblock In \emph{Proceedings of the Annual Meeting of the Association for Computational Linguistics}, pages 7881--7892.

\bibitem[{Si et~al.(2023)Si, Goyal, Wu, Zhao, Feng, Daum{\'e}~III, and Boyd-Graber}]{si2023large}
Chenglei Si, Navita Goyal, Sherry~Tongshuang Wu, Chen Zhao, Shi Feng, Hal Daum{\'e}~III, and Jordan Boyd-Graber. 2023.
\newblock Large language models help humans verify truthfulness--except when they are convincingly wrong.
\newblock \emph{arXiv preprint arXiv:2310.12558}.

\bibitem[{Stede(2008)}]{stede2008connective}
Manfred Stede. 2008.
\newblock \href {https://aclanthology.org/W08-2218} {Connective-based local coherence analysis: A lexicon for recognizing causal relationships}.
\newblock In \emph{Semantics in Text Processing. {STEP} Conference Proceedings}, pages 221--237.

\bibitem[{Tan et~al.(2023)Tan, Nguyen, Bensemann, Peng, Bao, Chen, Gahegan, and Witbrock}]{tan2023multi2claim}
Neset Tan, Trung Nguyen, Josh Bensemann, Alex Peng, Qiming Bao, Yang Chen, Mark Gahegan, and Michael Witbrock. 2023.
\newblock \href {https://doi.org/10.18653/v1/2023.eacl-main.194} {{M}ulti2{C}laim: Generating scientific claims from multi-choice questions for scientific fact-checking}.
\newblock In \emph{Proceedings of the Conference of the European Chapter of the Association for Computational Linguistics}, pages 2652--2664.

\bibitem[{Thomas(1973)}]{thomas1973practical}
Stephen~Naylor Thomas. 1973.
\newblock \emph{Practical Reasoning in Natural Language}.
\newblock Prentice-Hall.

\bibitem[{Thorne et~al.(2018)Thorne, Vlachos, Christodoulopoulos, and Mittal}]{thorne2018fever}
James Thorne, Andreas Vlachos, Christos Christodoulopoulos, and Arpit Mittal. 2018.
\newblock \href {https://doi.org/10.18653/v1/N18-1074} {{FEVER}: a large-scale dataset for fact extraction and {VER}ification}.
\newblock In \emph{Proceedings of the Conference of the North {A}merican Chapter of the Association for Computational Linguistics}, pages 809--819.

\bibitem[{Toulmin(2003)}]{toulmin2003uses}
Stephen~E Toulmin. 2003.
\newblock \emph{The uses of argument}.
\newblock Cambridge University Press.

\bibitem[{Walton(2013)}]{walton2013argumentation}
Douglas Walton. 2013.
\newblock \emph{Argumentation schemes for presumptive reasoning}.
\newblock Routledge.

\bibitem[{Walton et~al.(2008)Walton, Reed, and Macagno}]{walton2008argumentation}
Douglas Walton, Christopher Reed, and Fabrizio Macagno. 2008.
\newblock \emph{Argumentation schemes}.
\newblock Cambridge University Press.

\bibitem[{Wang et~al.(2023{\natexlab{a}})Wang, Li, and Ji}]{wang2022code4struct}
Xingyao Wang, Sha Li, and Heng Ji. 2023{\natexlab{a}}.
\newblock \href {https://doi.org/10.18653/v1/2023.acl-long.202} {{C}ode4{S}truct: Code generation for few-shot event structure prediction}.
\newblock In \emph{Proceedings of the Annual Meeting of the Association for Computational Linguistics}, pages 3640--3663.

\bibitem[{Wang et~al.(2023{\natexlab{b}})Wang, Le, Gotmare, Bui, Li, and Hoi}]{wang2023codet5+}
Yue Wang, Hung Le, Akhilesh Gotmare, Nghi Bui, Junnan Li, and Steven Hoi. 2023{\natexlab{b}}.
\newblock \href {https://doi.org/10.18653/v1/2023.emnlp-main.68} {{C}ode{T}5+: Open code large language models for code understanding and generation}.
\newblock In \emph{Proceedings of the Conference on Empirical Methods in Natural Language Processing}, pages 1069--1088.

\bibitem[{Wang et~al.(2021)Wang, Wang, Joty, and Hoi}]{wang2021codet5}
Yue Wang, Weishi Wang, Shafiq Joty, and Steven~C.H. Hoi. 2021.
\newblock \href {https://doi.org/10.18653/v1/2021.emnlp-main.685} {{C}ode{T}5: Identifier-aware unified pre-trained encoder-decoder models for code understanding and generation}.
\newblock In \emph{Proceedings of the Conference on Empirical Methods in Natural Language Processing}, pages 8696--8708.

\bibitem[{Wei et~al.(2022)Wei, Wang, Schuurmans, Bosma, Xia, Chi, Le, Zhou et~al.}]{wei2022chain}
Jason Wei, Xuezhi Wang, Dale Schuurmans, Maarten Bosma, Fei Xia, Ed~Chi, Quoc~V Le, Denny Zhou, et~al. 2022.
\newblock Chain-of-thought prompting elicits reasoning in large language models.
\newblock \emph{Advances in Neural Information Processing Systems}, 35:24824--24837.

\bibitem[{Willig et~al.(2023)Willig, Ze{\v{c}}evi{\'c}, Dhami, and Kersting}]{willig2022probing}
Moritz Willig, Matej Ze{\v{c}}evi{\'c}, Devendra~Singh Dhami, and Kristian Kersting. 2023.
\newblock \href {https://openreview.net/forum?id=UPwzqPOs4-} {Probing for correlations of causal facts: Large language models and causality}.

\bibitem[{Yao et~al.(2023)Yao, Zhao, Yu, Du, Shafran, Narasimhan, and Cao}]{yao2022react}
Shunyu Yao, Jeffrey Zhao, Dian Yu, Nan Du, Izhak Shafran, Karthik~R Narasimhan, and Yuan Cao. 2023.
\newblock \href {https://openreview.net/forum?id=WE_vluYUL-X} {React: Synergizing reasoning and acting in language models}.
\newblock In \emph{Proceedings of the International Conference on Learning Representations}.

\bibitem[{Yu et~al.(2019)Yu, Li, and Wang}]{yu2019detecting}
Bei Yu, Yingya Li, and Jun Wang. 2019.
\newblock \href {https://doi.org/10.18653/v1/D19-1473} {Detecting causal language use in science findings}.
\newblock In \emph{Proceedings of the Conference on Empirical Methods in Natural Language Processing}, pages 4664--4674.

\bibitem[{Zhang et~al.(2023)Zhang, Bauer, Bennett, Gao, Gong, Hilmkil, Jennings, Ma, Minka, Pawlowski et~al.}]{zhang2023understanding}
Cheng Zhang, Stefan Bauer, Paul Bennett, Jiangfeng Gao, Wenbo Gong, Agrin Hilmkil, Joel Jennings, Chao Ma, Tom Minka, Nick Pawlowski, et~al. 2023.
\newblock Understanding causality with large language models: Feasibility and opportunities.
\newblock \emph{arXiv preprint arXiv:2304.05524}.

\bibitem[{Zhang et~al.(2024)Zhang, Zhang, and Zhou}]{causalwalk}
Congzhi Zhang, Linhai Zhang, and Deyu Zhou. 2024.
\newblock \href {https://doi.org/10.1609/aaai.v38i17.29925} {Causal walk: Debiasing multi-hop fact verification with front-door adjustment}.
\newblock In \emph{Proceedings of the AAAI Conference on Artificial Intelligence}.

\bibitem[{Zhao et~al.(2020)Zhao, Xiong, Rosset, Song, Bennett, and Tiwary}]{zhao2019transformer}
Chen Zhao, Chenyan Xiong, Corby Rosset, Xia Song, Paul~N. Bennett, and Saurabh Tiwary. 2020.
\newblock \href {https://api.semanticscholar.org/CorpusID:212745235} {Transformer-xh: Multi-evidence reasoning with extra hop attention}.
\newblock In \emph{Proceedings of the International Conference on Learning Representations}.

\bibitem[{Zhu et~al.(2023)Zhu, Si, Zhao, Zhu, Zhou, and He}]{zhu2023explain}
Yingjie Zhu, Jiasheng Si, Yibo Zhao, Haiyang Zhu, Deyu Zhou, and Yulan He. 2023.
\newblock \href {https://doi.org/10.18653/v1/2023.emnlp-main.826} {{EXPLAIN}, {EDIT}, {GENERATE}: Rationale-sensitive counterfactual data augmentation for multi-hop fact verification}.
\newblock In \emph{Proceedings of the Conference on Empirical Methods in Natural Language Processing}, pages 13377--13392.

\end{thebibliography}

\newpage

\appendix

\newpage

\section{Experiment Details}
\label{Experiment Details}
In the fine-tuning process for Discriminative Models, we choose BERT-base-uncased, Transformer-XH (using BERT-base-uncased as the backbone), and UniXcoder-base. The batch size is set to 8, and we utilize the AdamW optimizer with a learning rate of 5e-6.

We select CodeT5-base and CodeT5+ (0.7B) for fine-tuning generation models, with a batch size of 4 and a learning rate of 1e-5. Additionally, we specify a maximum input length of 512 and a maximum generation length of 400.

Regarding Large Language Models (LLMs), our choices are gpt-4-1106-preview and ChatGPT-turbo, with a temperature setting of 0.1.

\section{Ablation Experiment}
\label{ablation experiment}
\paragraph{Number of Irrelevant Evidence}
We perform experiments on Task 4 with different average amounts of irrelevant evidence. The findings are summarized in Table~\ref{distractor}, suggesting that task difficulty increases slightly as the number of irrelevant evidence increases.

\paragraph{Number of Reasoning Steps}
We group the results of Task 4 according to the number of reasoning steps in the gold argument structures. The outcomes are outlined in Table~\ref{argument steps broke}, demonstrating a significant decrease in scores as the number of steps increases.

\paragraph{Further Evaluation}
We refer to the evaluation metrics from \textsc{EntailmentBank} and make slight modifications to further evaluate the argument structures generated in Task 3 and Task 4. The specific results are shown in the Table~\ref{entaillment result task34}.

• Leaf Nodes (F1, AllCorrect): Does the predicted argument structure use the correct leaf evidence? We compute an F1 score by comparing predicted leaf evidence to golden leaf evidence. The “AllCorrect” score is 1 if all nodes are identified correctly (F1=1.0), 0 otherwise.

• Steps (F1, AllCorrect): Are the individual reasoning steps structurally correct? As each intermediate node represents (the conclusion of) a single step, the step is considered structurally correct (score 1) if it perfectly matches the gold, 0 otherwise. We then measure F1 comparing all steps in the two trees. Then AllCorrect=1 if F1=1.0, 0 otherwise.

• Intermediates (F1, AllCorrect): Are the intermediate conclusions correct? For comparing gold and generated conclusions. F1 is computed using the number of aligned, correct intermediates wrt. the number of gold/predicted intermediates. AllCorrect=1 if F1=1, otherwise 0.

Based on the experimental results, the performance of the fine-tuned model significantly surpasses that of large language models (LLMs). From the Leaves metric, it is evident that the model can identify most of the evidence related to verification. However, the last two metrics indicate that the model encounters difficulties in linking these pieces of evidence to form an argumentative structure.

\paragraph{Pipeline Setup}
We also design an experiment in pipeline mode, using both CodeT5 and CodeT5+ as baseline models. In the first stage, models output argument structures, and in the second stage, they produce the final classification results. The experimental results shown in Table~\ref{pipeline} demonstrate that CodeT5+ generates better argument structures, and achieves better classification performance.

\begin{figure}[h]
	\centering
	\resizebox{0.8\linewidth}{!}{
		\includegraphics{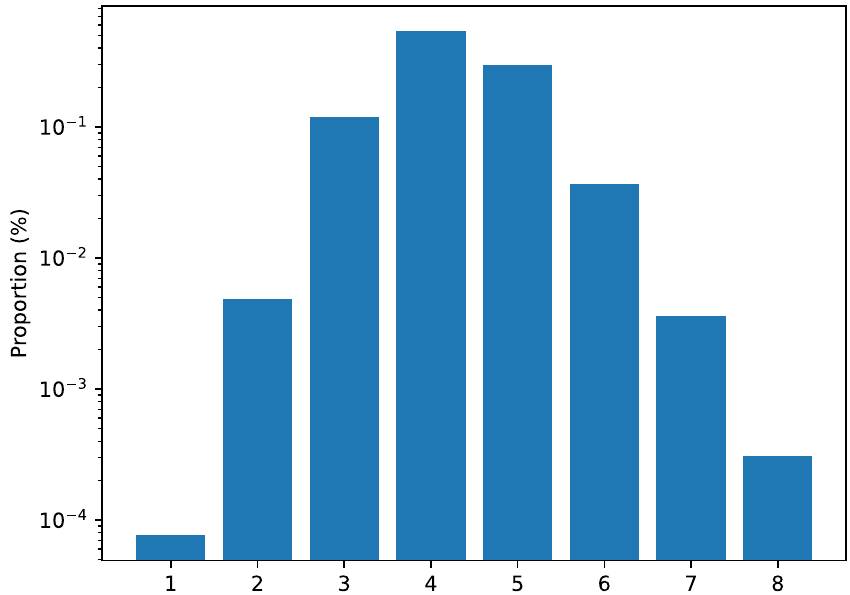}
	}
	\caption{Histogram of useful evidence number in the C\textsc{heck}W\textsc{hy}}
	\label{evidence number}
\end{figure}

\begin{figure}[h]
	\centering
	\resizebox{0.8\linewidth}{!}{
		\includegraphics{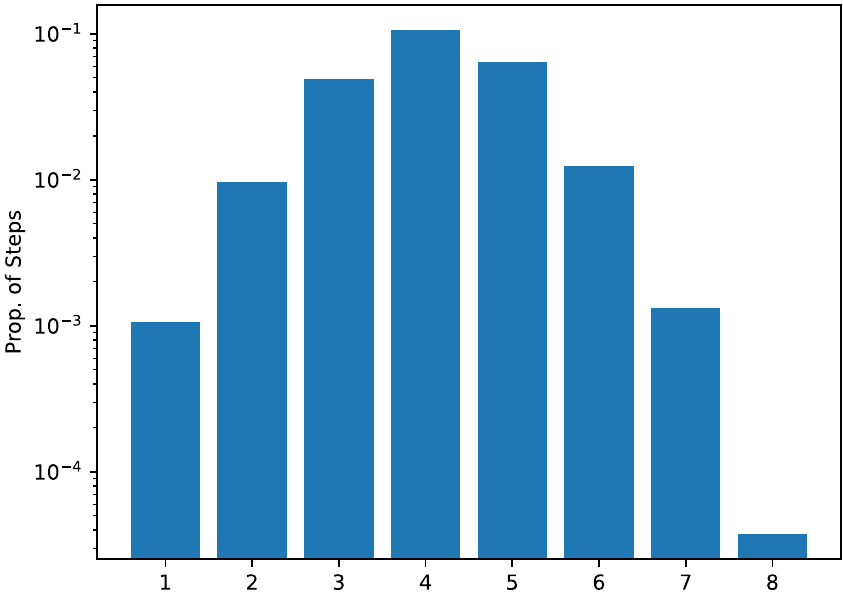}
	}
	\caption{Histogram of argument steps in the C\textsc{heck}W\textsc{hy}}
	\label{argument steps}
\end{figure}

\begin{table}[h]
	\centering
	\resizebox{0.85\linewidth}{!}{
		\begin{tabular}{ccccc}
			\toprule[1pt]
			\small
			\textbf{Number of Irrelevant Evidence} & \textbf{Acc} & \textbf{F1} & \textbf{SS} & \textbf{ES} \\
			\midrule[0.5pt]
			6                                      & 67.6         & 66.8        & 72.3        & 1.3         \\
			8                                      & 69.3         & 69.1        & 75.7        & 2.5         \\
			10                                     & 65.4         & 63.9        & 71.9        & 1.0         \\
			12                                     & 62.3         & 60.0        & 71.0        & 1.5         \\
			\bottomrule[1pt]
		\end{tabular}}
	\caption{\label{distractor} The performance of CodeT5 on Task 4 involving different numbers of irrelevant evidence.}
\end{table}

\begin{table}[h]
	\centering
	\resizebox{1\linewidth}{!}{
		\begin{tabular}{cccccccc}
			\toprule[1.5pt]
			\textbf{Number of Steps} & \textbf{Number of instances} & \textbf{Acc} & \textbf{F1} & \textbf{SS} & \textbf{ES} \\

			\midrule[0.5pt]
			1                        & 40                           & 80.0           & 80.0          & 78.6        & 8.6         \\
			2                        & 273                          & 74.3         & 72.0          & 77.4        & 2.6         \\
			3                        & 734                          & 68.9         & 66.4        & 73.3        & 0.7         \\
			4                        & 464                          & 65.7         & 65.4        & 70.4        & 0.1         \\
			5                        & 77                           & 64.9         & 59.6        & 64.6        & 0.3         \\
			$\geq$ 6                 & 12                           & 58.3         & 49.6        & 62.7        & 0.0           \\
			\bottomrule[1.5pt]
		\end{tabular}}
	\caption{Results on Task 4 with the varying number of argument steps in the gold structure.}
	\label{argument steps broke}
\end{table}

\begin{table}[h]
	\centering
	\small
	\resizebox{0.7\linewidth}{!}{
		\begin{tabular}{ccccc}
			\toprule[1pt]
			\textbf{Model} & \textbf{Acc} & \textbf{F1} & \textbf{SS} & \textbf{ES} \\

			\midrule[0.5pt]
			CodeT5         & 71.1         & 70.7        & 72.9        & 1.4         \\
			CodeT5+        & 75.9         & 76.1        & 76.3        & 2.6         \\
			\bottomrule[1pt]
		\end{tabular}}
	\caption{Results on Task 4 with the pipeline setup.}
	\label{pipeline}
\end{table}

\begin{table*}[t]
	\small
	\centering
	\resizebox{0.8\linewidth}{!}{
		\begin{tabular}{cccccccccc}
			\toprule[1pt]
			% \multicolumn{2}{l}{\multirow{2.5}{*}{\textbf{Model}}} & \multicolumn{2}{c}{\textbf{Automatic}} & \multicolumn{4}{c}{\textbf{Human}} & \multicolumn{2}{c}{\textbf{Verification}} \\
			\multicolumn{2}{c}{\multirow{2.5}{*}{\textbf{Model}}} & \multicolumn{2}{c}{\textbf{Automatic}}                      & \multicolumn{2}{c}{\textbf{Human}} & \multicolumn{2}{c}{\textbf{Verification}}                                                                \\
			\cmidrule(lr){3-4}\cmidrule(lr){5-6}\cmidrule(lr){7-8}
			\multicolumn{2}{c}{}
			                                                      & F1                                                          & AllCorrect                         & F1                                        & AllCorrect    & F1            & AllCorrect                   \\
			\midrule[0.7pt]
			\multirow{7}{*}{\textbf{{Task 3}}}                    & \multicolumn{7}{l}{\textit{\textbf{Generative Models}}}                                                                                                                                                     \\
			                                                      & CodeT5 (FT)                                                 & \textbf{85.6}                      & \textbf{60.8}                             & \textbf{20.7} & \textbf{ 3.4} & 36.1          & 4.4          \\
			                                                      & CodeT5+ (FT)                                                & 84.4                               & 59.3                                      & 20.5          & \textbf{3.4}  & \textbf{36.6} & \textbf{6.3} \\
			\cmidrule[0.2pt](r){2-10}
			                                                      & \multicolumn{9}{l}{\textit{\textbf{Large Language Models}}}                                                                                                                                                 \\
			                                                      & GPT-4 (CoT)                                                 & 51.2                               & 12.4                                      & 9.1           & 0.2           & 14.2          & 1.42         \\
			                                                      & ChatGPT (CoT)                                               & 43.2                               & 16.4                                      & 10.0          & 0.0             & 14.0          & 0.6          \\

			\midrule[0.7pt]
			\multirow{7}{*}{\textbf{{Task 4}}}                    & \multicolumn{7}{l}{\textbf{\textit{Generative Models}}}                                                                                                                                                     \\
			                                                      & CodeT5 (FT)                                                 & \textbf{81.1}                      & 46.3                                      & 9.2           & 0.06          & \textbf{12.2} & 0.06         \\
			                                                      & CodeT5+ (FT)                                                & 77.8                               & 42.6                                      & 15.4          & \textbf{8.7}  & 9.4           & 0.06         \\
			\cmidrule[0.2pt](r){2-10}
			                                                      & \multicolumn{9}{l}{\textbf{\textit{Large Language Models}}}                                                                                                                                                 \\
			                                                      & GPT-4 (CoT)                                                 & 79.7                               & \textbf{53.4}                             & \textbf{16.4} & 7.2           & 8.16          & 0.2          \\
			                                                      & ChatGPT (CoT)                                               & 61.3                               & 32.7                                      & 15.5          & 3.7           & 7.6           & \textbf{0.4} \\

			\bottomrule[1pt]
		\end{tabular}}
	\caption{The results evaluated using the metrics from the \textsc{EntailmentBank}~\cite{dalvi2021explaining}. }
	\label{entaillment result task34}
\end{table*}

\begin{table*}[h]
	\centering
	\resizebox{.8\linewidth}{!}{
		\begin{tabular}{lccc}
			\toprule[1pt]
			\textbf{Dataset}                    & \textbf{Reasoning Structure} & \textbf{Complicated Multi-hop} & \textbf{Causal Verification} \\ \midrule[0.5pt]
			FEVEROUS                            & \XSolidBrush                 & \CheckmarkBold                 & \XSolidBrush                 \\
			FEVER                               & \XSolidBrush                 & \XSolidBrush                   & \XSolidBrush                 \\
			\textsc{HoVer}                      & \XSolidBrush                 & \CheckmarkBold                 & \XSolidBrush                 \\ \midrule[0.5pt]
			\textbf{C\textsc{heck}W\textsc{hy}} & \CheckmarkBold
			                                    & \CheckmarkBold               & \CheckmarkBold                                                \\ \bottomrule[1pt]
		\end{tabular}}
	\caption{\label{table comparison} Comparison of C\textsc{heck}W\textsc{hy} with previous Fact Verification datasets}
\end{table*}

\section{Detailed Description of the Annotation Process}
\label{Detailed Description of the Annotation Process}
\paragraph{Source of Annotator}
To ensure the quality of our complex annotation process, we do not adopt CrowdSource platforms, even though more CrowdWorkers can be employed. We hire 20 university students with experience in NLP, especially those majoring in logical reasoning or argument mining. Before the formal annotation, we draft detailed guidelines, set up the annotation platform, and conduct three rounds of thorough training and one round of testing. In this way, 13 students pass the exam and are retained. Furthermore, we split these students into two groups, where 1 PhD student and 8 postgraduate students perform the Editing, and 1 PhD student and 3 postgraduate students perform the Filtering. Moreover, we apply a strict quality control procedure during our annotation process. In addition, authors conduct casual inspections during the annotation process and rate each annotator, with annotators who receive low scores undergoing additional training.

\paragraph{Annotation Cost}
The annotating time for each sample ranges from 8 to 30 minutes, averaging 15 minutes per sample. After annotation, we follow local labor laws, and each annotator is paid \$3.05 per hour.

Due to the complex argument structure within each instance, the analysis of argument structure type and evaluation of prediction is quite time-consuming and labor-sensitive, especially facing complex structures. The average time for analyzing argument structure type is more than 20 minutes per instance, and the average time of evaluation on prediction ranges from 15 minutes to 20 minutes per instance, depending on different baseline models.

\paragraph{Quality Contol}
We implemented several quality control measures to minimize the possibility of annotators cutting corners and to ensure the quality of the data:

\noindent $\bullet$ \textbf{Editing} Given each instance, two annotators are required to perform manual editing of the Claim and Evidence and to resolve any errors identified by the third annotator. The final result is chosen by voting from all three annotators. Due to the complexity of the argument structure and its highly time-consuming nature, we don't apply this procedure to the editing of the argument structure. %TODO 疑问

\noindent $\bullet$ \textbf{Filtering} We apply a strict quality control procedure by filtering each instance with three anonymous annotators, wherein two annotators are required to filter the instance, and the third annotator (the Ph.D. student) serves as the supervisor. The decision of each instance is made by all three annotators. Moreover, the data transfer is completed by our annotation platform. According to our statistics, the ratio of "all-pass" is acceptable. A display of an example modified by humans is shown in Table~\ref{modifi}.

\noindent $\bullet$ \textbf{Automatic Check} We have implemented a grammar check mechanism(e.g. the argument graph must be a connected graph, etc.) to verify the legality of the annotated argument structures. Before annotators can submit their results, they must pass this check.

\noindent $\bullet$ \textbf{Human Check} The authors conduct a random casual inspection of the edited data and rate each annotator. Moreover, since we have separated the annotators into two groups at different stages if the annotators in Filtering find low-quality edited instances, the annotators in Editing will receive a low rating.

\section{Annotation Interface}
\label{Annotation Interface}
\begin{figure*}[h]
	\centering
	\includegraphics[width=0.8\textwidth]{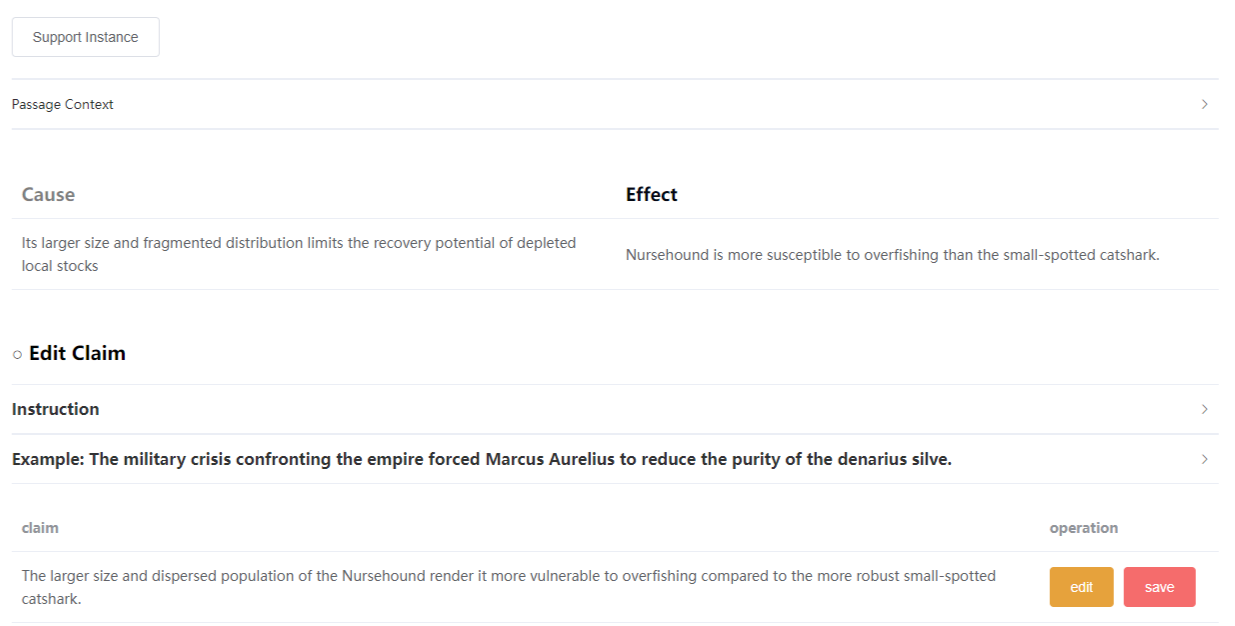}
	\vspace{-2ex}
	\caption{Claim Annotation Interface}
	\label{Annotation Interface 1}
\end{figure*}

\begin{figure*}[h]
	\centering
	\includegraphics[width=0.8\textwidth]{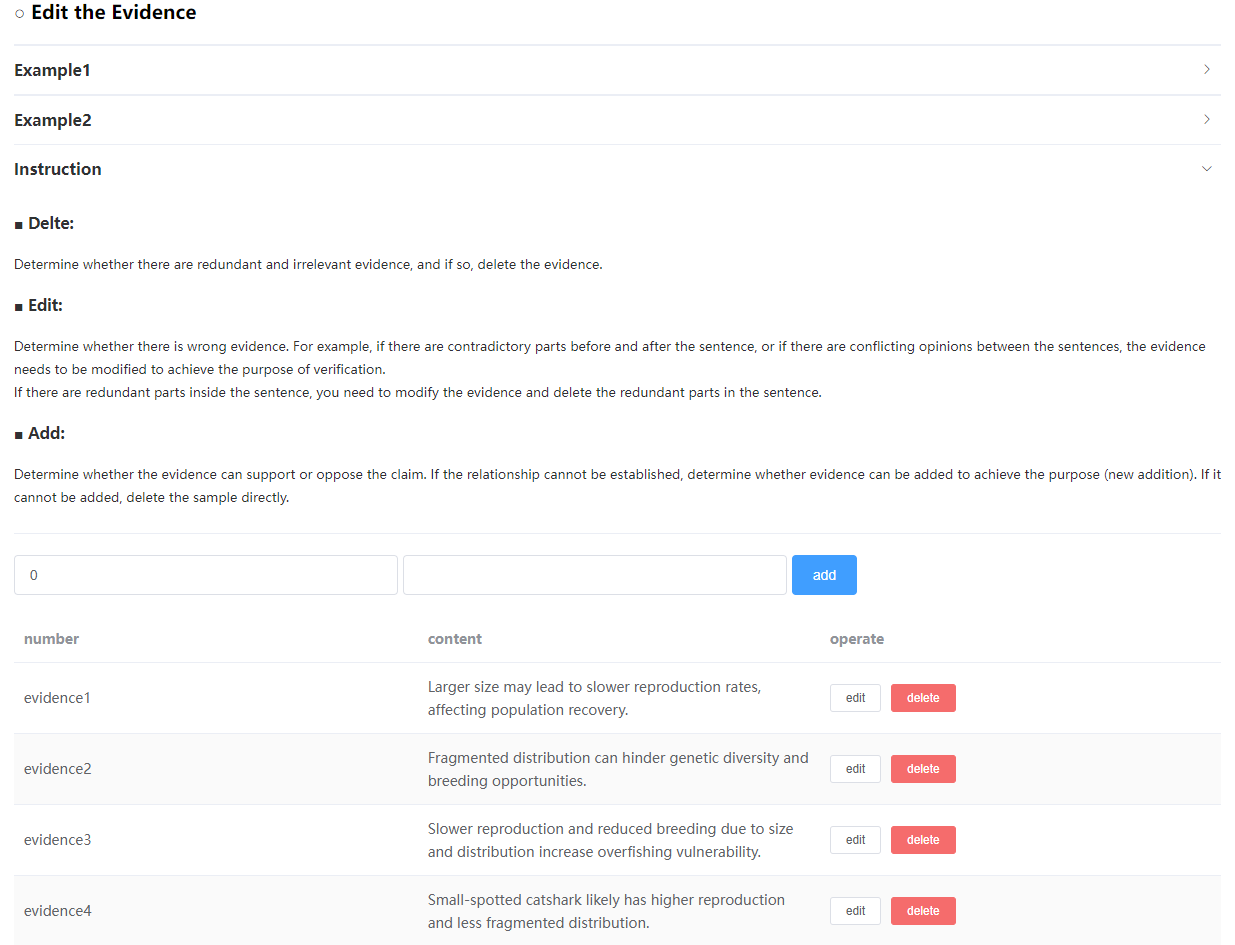}
	\vspace{-2ex}
	\caption{Evidence Annotation Interface}
	\label{Annotation Interface 2}
\end{figure*}

\begin{figure*}[h]
	\centering
	\includegraphics[width=0.8\textwidth]{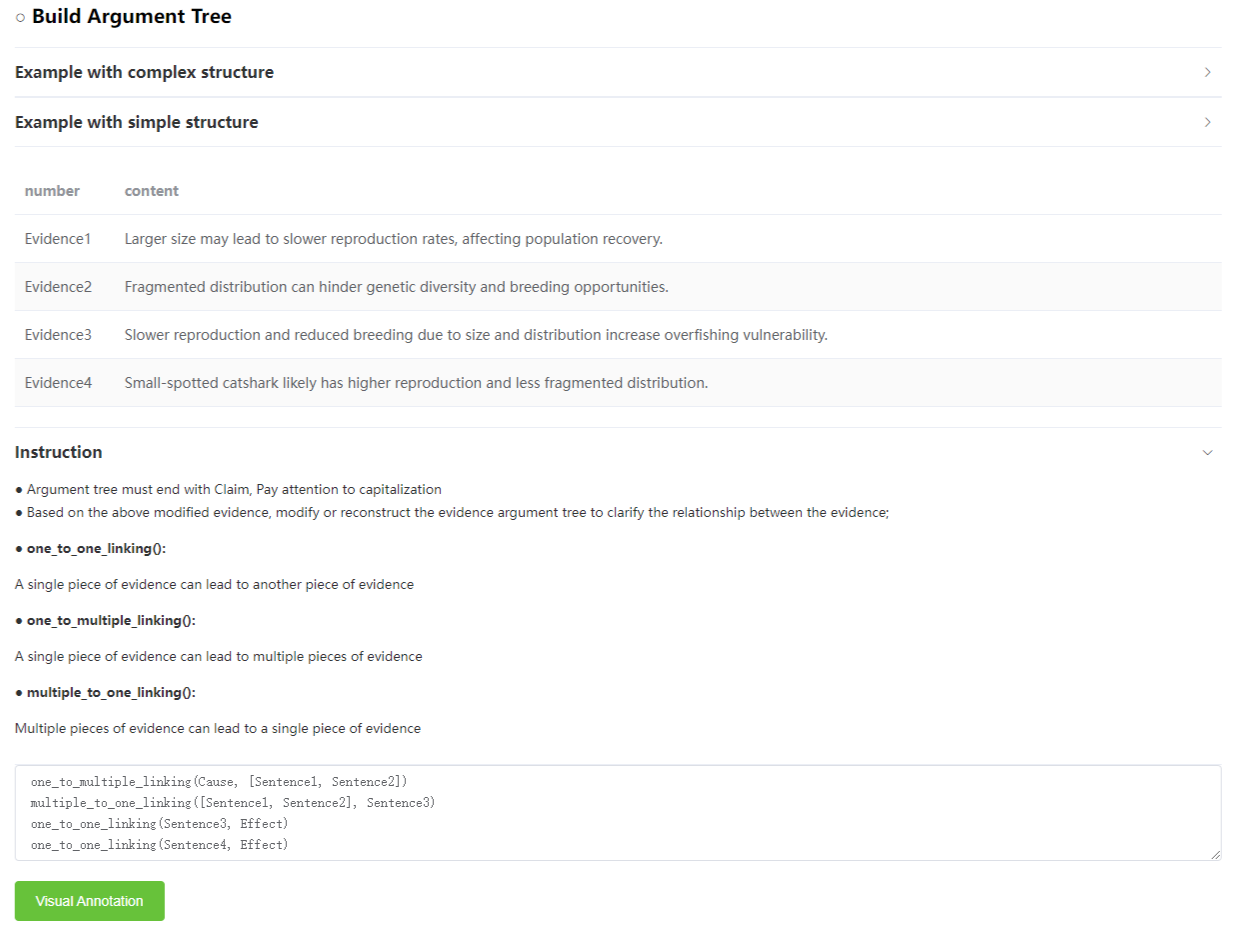}
	\vspace{-2ex}
	\caption{Argument Structure Annotation Interface}
	\label{Annotation Interface 3}
\end{figure*}

\begin{figure*}[h]
	\centering
	\includegraphics[width=0.8\textwidth]{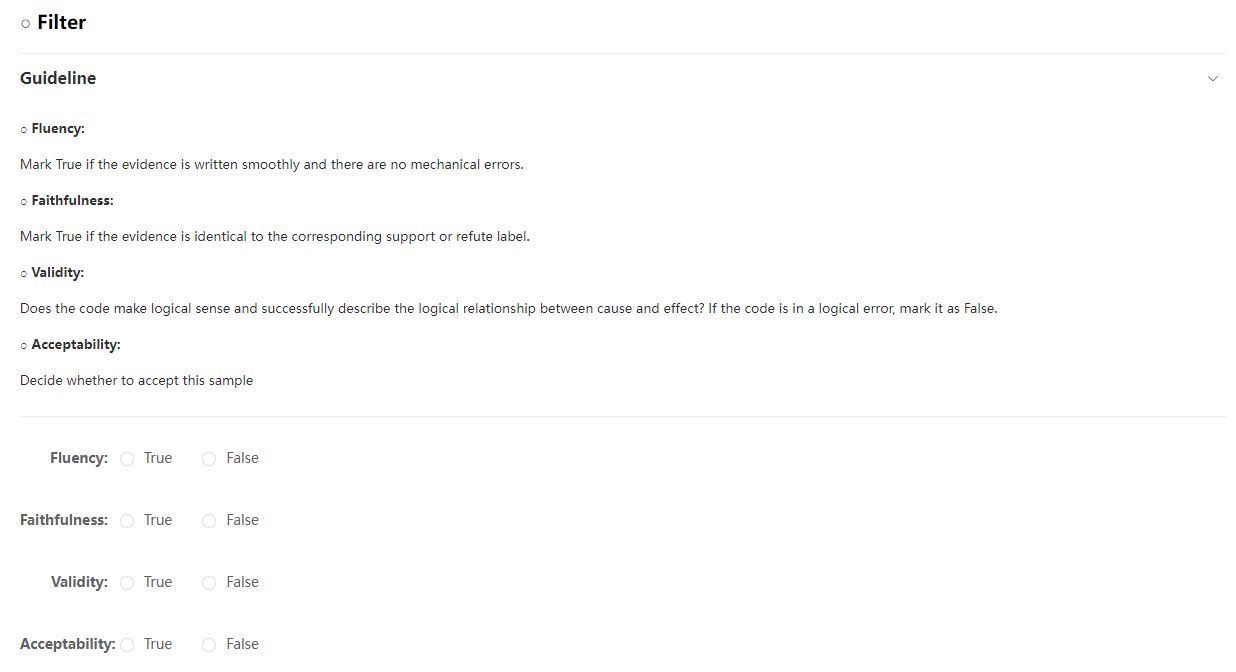}
	\vspace{-2ex}
	\caption{Filter Annotation Interface}
	\label{Annotation Interface 4}
\end{figure*}

Figure~\ref{Annotation Interface 1} to Figure~\ref{Annotation Interface 4} illustrate the interfaces utilized for claim annotation, evidence annotation, argument map annotation, and the final review stage. These figures include comprehensive instructions, as well as contextual information for cause-effect pairs, to aid annotators in their understanding. Additionally, examples for each annotation stage are provided to enhance annotator proficiency.

\section{Generation Prompts}
\label{Generation Prompts}
The prompts utilized during the data collection stage are depicted in Table~\ref{claim prompt} to Table~\ref{code prompt}. Meanwhile, the prompts for experiments can be found in Table~\ref{task1 prompt} to Table~\ref{task4 prompt}.

\begin{table*}[]
	\centering
	\resizebox{0.9\linewidth}{!}{
		\renewcommand\arraystretch{1}
		\begin{tabular}{l}
			\toprule[1pt]

			\textbf{Claim}: The existence of an ancient lake thousands of years ago has led to the flat geography of South Jordan today.                             \\ \hline
			\textbf{GPT-4}: SUPPORT Evidence                                                                                                                         \\ \hline
			G1: Ancient lakes, when dried up, leave behind sediment that is evenly spread out, resulting in flat land.                                               \\
			G2: Geological evidence suggests that South Jordan was once covered by an ancient lake that left sediment deposits.                                      \\
			G3: The sediment from the ancient lake in South Jordan has, over thousands of years, compacted and created a flat landscape.                             \\
			G4: No significant geological events have occurred to disrupt the flatness of the ancient lake bed in South Jordan.                                      \\
			G5: The ancient lake's flat bed is the primary reason for South Jordan's flatness, as no other events have altered the landscape.                        \\ \hline
			\textbf{GPT4}: Argument Structure                                                                                                                        \\ \hline
			G2 $\rightarrow$  (G1, G3) $\rightarrow$  G5                                                                                                             \\
			(G4, G5) $\rightarrow$  Claim                                                                                                                            \\ \hline
			\textbf{Human}: SUPPORT Evidence                                                                                                                         \\ \hline
			E1: Ancient lakes, when dried up, leave behind sediment that is evenly spread out\sout{,resulting in flat land}.                                         \\
			E2: Geological evidence suggests that South Jordan was once covered by an ancient lake\sout{that left sediment deposits}.                                \\
			E3: In the South Jordan area, dried-up lakes leave behind sediment that is evenly distributed.                                                           \\
			E4: Sediments from the ancient lake in South Jordan have been \textbf{compacted} over thousands of years\sout{, compacted and created a flat landscape}. \\
			E5: No significant geological events have occurred to disrupt the \sout{flatness} \textbf{landscape} of the ancient lake bed in South Jordan.            \\
			\sout{G5: The ancient lake's flat bed is the primary reason for South Jordan's flatness, as no other events have altered the landscape.}                 \\
			E6: \textbf{The compacted sediment may eventually form a flat landscape.}                                                                                \\ \hline
			\textbf{Human}: Argument Structure                                                                                                                       \\ \hline
			(E1, E2) $\rightarrow$ E3 $\rightarrow$ E4 $\rightarrow$ E6                                                                                              \\
			(E5, E6) $\rightarrow$ Claim                                                                                                                             \\

			\bottomrule[1pt]
		\end{tabular}
	}
	\caption{A comparison example between texts written by GPT-4 and human modifications.}
	\label{modifi}
\end{table*}

\begin{figure*}[h]
	\centering
	\includegraphics[width=0.8\textwidth]{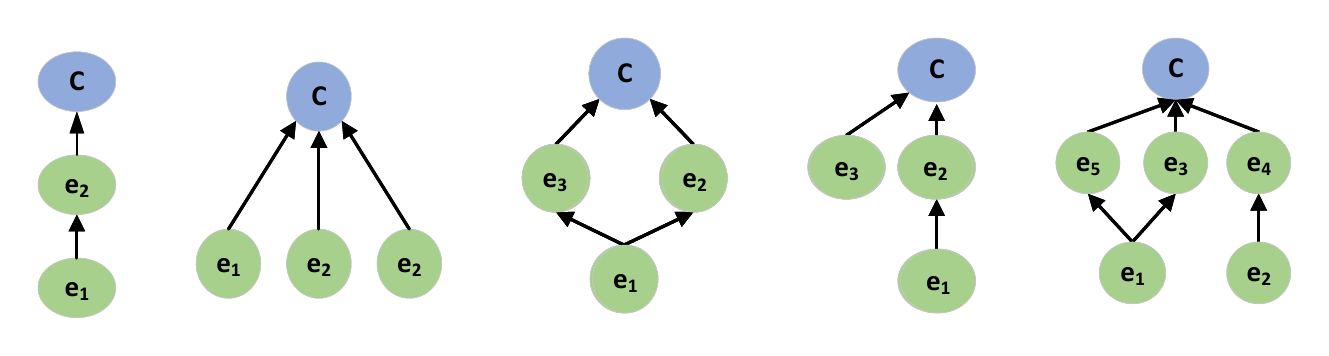}
	\vspace{-2ex}
	\caption{Types of Argument Structure}
	\label{Types of argument diagrams}
\end{figure*}

\begin{figure*}[h]
	\centering
	\includegraphics[width=13cm]{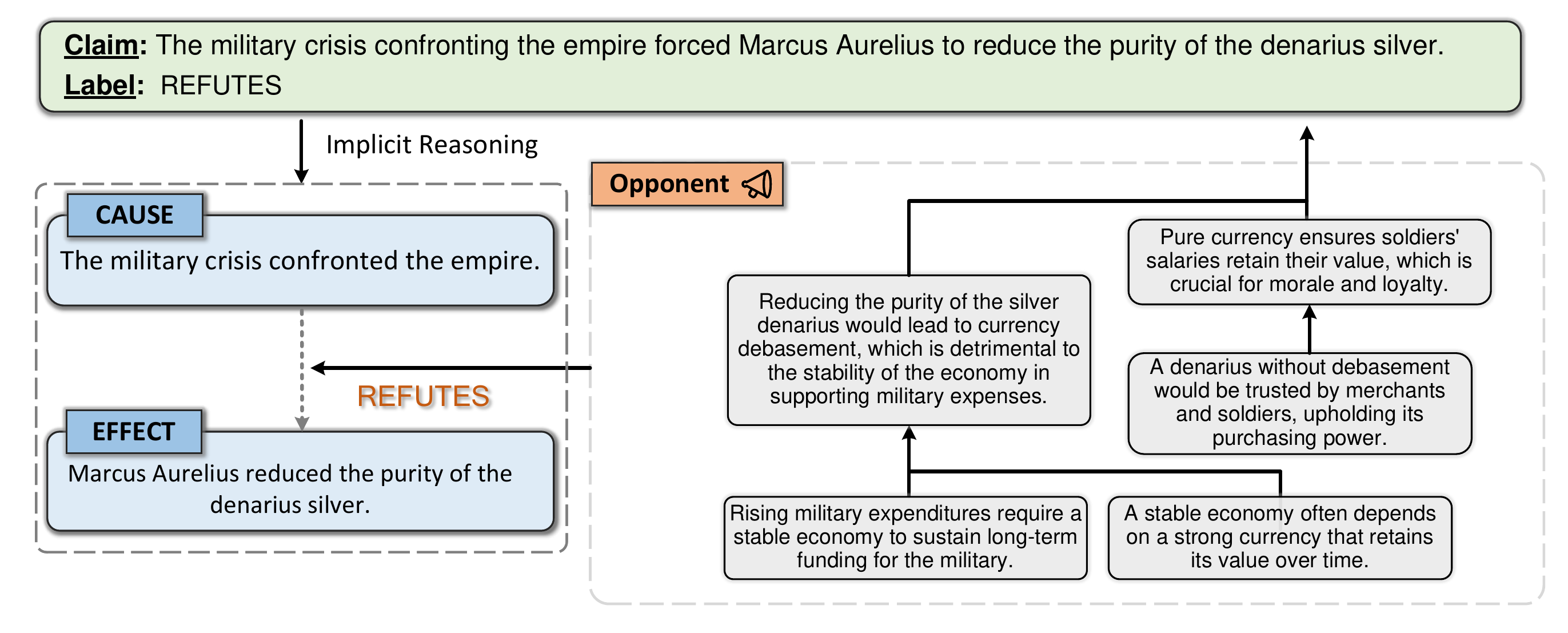}
	\vspace{-1ex}
	\caption{An entry with \textit{REFUTES} labels from C\textsc{heck}W\textsc{hy}}
	\label{REFUTES example}
\end{figure*}

\begin{figure*}[h]
	\centering
	\includegraphics[width=13cm]{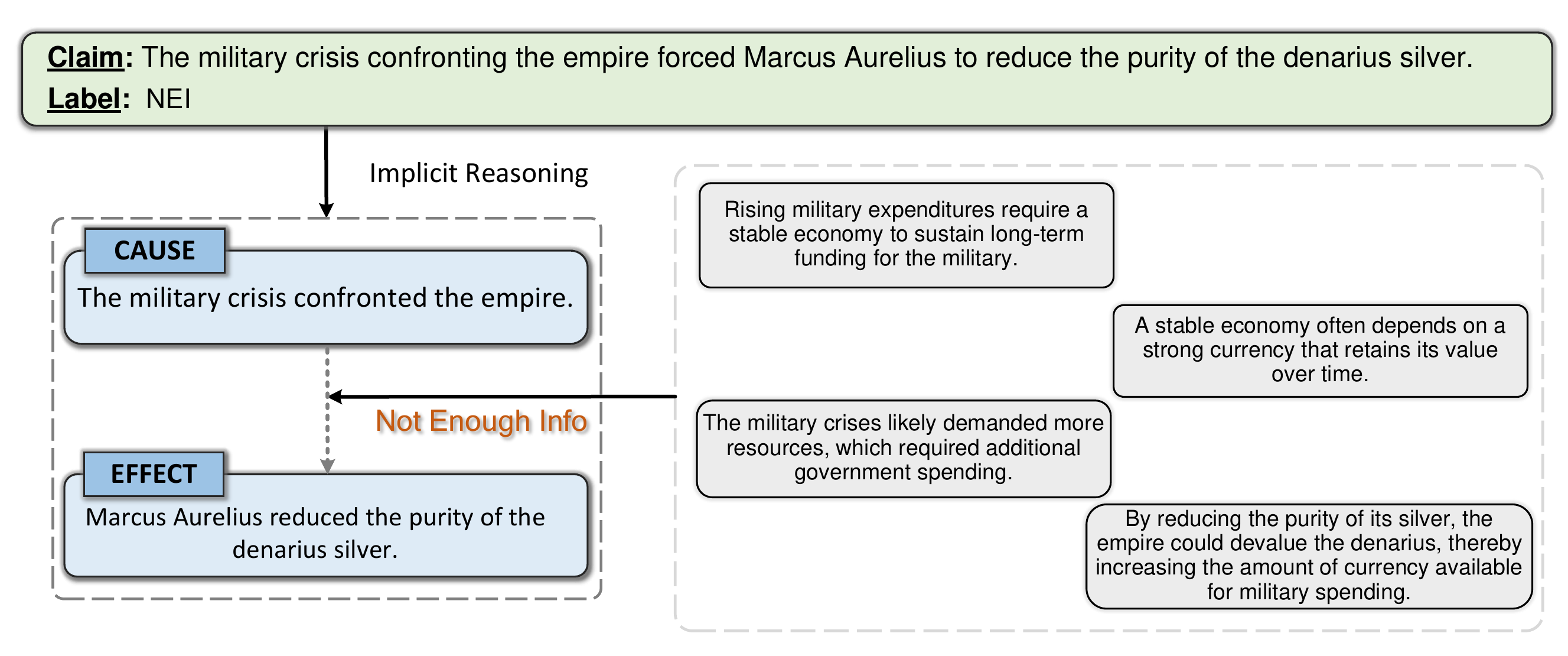}
	\vspace{-1ex}
	\caption{An entry with \textit{NEI} labels from C\textsc{heck}W\textsc{hy}}
	\label{NEI example}
\end{figure*}

\begin{table*}[h]
	\centering
	\resizebox{0.8\linewidth}{!}{
		\begin{tabularx}{15.5cm}{X}
			\toprule[1pt]
			I am an excellent linguist. Your task is to generate a refined claim by synthesizing the content of the \{Cause\} and the \{Effect\}.                  \\
			Be sure to use creative and diverse ways to generate the \{Claim\}.                                                                                    \\
			You will be provided with                                                                                                                              \\
			\{\textbf{Cause}\} is the reason why the \{Effect\} is established;                                                                                    \\
			\{\textbf{Effect}\} is the consequence of \{Cause\}.                                                                                                   \\
			\textbf{Note:}                                                                                                                                         \\
			No matter whether this pair of \{Cause\} and \{effect\} is correct or not, you must synthesize a concise and clean \{Claim\} with causality contained. \\
			Do not induce any extra information in your generation except what we provided.                                                                        \\
			\textbf{Example:}                                                                                                                                      \\
			\textbf{Cause}: Powerranger's film market is in the doldrums.                                                                                          \\
			\textbf{Effect}: It's not possible that the Power Rangers series will get a sequel.                                                                    \\
			\textbf{Claim}: The lackluster performance of the Power Rangers film in the market has plunged its sequel prospects into uncertainty.                  \\
			...                                                                                                                                                    \\
			\bottomrule[1pt]
		\end{tabularx}}
	\caption{Prompt for Claim Generation}
	\label{claim prompt}
\end{table*}

\begin{table*}[h]
	\centering
	\resizebox{0.8\linewidth}{!}{
		\begin{tabularx}{15.5cm}{X}
			\toprule[1pt]
			I am an excellent logician. Your task is to generate persuasive factual evidence by expanding the \{anchor explanation\} to explain why a \{cause\} can lead to the \{effect\} in a step-by-step manner.                                                                                                                                                                       \\
			The \{anchor explanation\} provides the basic explanation you should refer to and generate around.                                                                                                                                                                                                                                                                             \\
			Solving this task with multiple steps, each step includes interleaving \{Thought\}, \{Action\}, \{Inference\}.                                                                                                                                                                                                                                                                 \\
			\{\textbf{Thought}\} is the analysis of the previous step by combining the information of \{cause\} \{effect\} pair, the generated entailment \{Inference\} in previous steps, and the \{anchor explanation\};                                                                                                                                                                 \\
			\{\textbf{Thought}\} is also to infer which \{Action\} should be taken next to reach the \{effect\} from the current step;                                                                                                                                                                                                                                                     \\
			\{\textbf{Inference}\} is the result of the current step when taking the \{Action\};                                                                                                                                                                                                                                                                                           \\
			\{\textbf{Action}\} is the solution you can operate based on the \{Thought\} in current step, and consists of 3 types:                                                                                                                                                                                                                                                         \\
			(1) \{\textbf{Further Reasoning}\}: infer directly using the content of the previous \{Inference\}, \{cause\}, and the \{anchor explanation\}.                                                                                                                                                                                                                                 \\
			(2) \{\textbf{Add new condition}\}: provide new factual condition to support the \{cause\} \{efect\} pair. Do this step only when you need context information that has not been provided in the \{cause\} \{effect\} pair and the \{anchor explanation\}, to make the inference logically and reasonably.                                                                     \\
			(3) \{\textbf{Finish}\}: summary the final results based on the \{Inference\}.                                                                                                                                                                                                                                                                                                 \\
			\\
			(Don't simply list facts, but reason in a detailed and step-by-step manner.)                                                                                                                                                                                                                                                                                                   \\
			\\
			\textbf{Example:}                                                                                                                                                                                                                                                                                                                                                              \\
			\textbf{cause}: Virtue of compassion for all living things in Buddhism.                                                                                                                                                                                                                                                                                                        \\
			\textbf{effect}: Qisong argued that Buddhist ethics were superior to Confucian ethics.                                                                                                                                                                                                                                                                                         \\
			anchor explanation: Confucian ethics do not dictate compassion for all living things.                                                                                                                                                                                                                                                                                          \\
			Building support reasoning process:                                                                                                                                                                                                                                                                                                                                            \\
			Thought1: By referring to the anchor explanation and the Cause, the difference between Buddhism and Confucianism towards compassion for all living things leads to the Qisong arguing that Buddhist ethics are superior to Confucian ethics. To support the cause-effect pair,  I need to understand in detail what would make Qisong argue Buddhism against Confucian ethics. \\
			\textbf{Action1: Further Reasoning}                                                                                                                                                                                                                                                                                                                                            \\
			...                                                                                                                                                                                                                                                                                                                                                                            \\
			\textbf{Thought5}: Based on the above reasoning, I can conclude that the final effect is established: Qisong argued that Buddhist ethics were superior to Confucian ethics.                                                                                                                                                                                                    \\
			\textbf{Action5: Finish}                                                                                                                                                                                                                                                                                                                                                       \\
			\textbf{Summary}: Based on the Inference1, Inference2, Inference3, and Inference4, the cause-effect pair is supported.                                                                                                                                                                                                                                                         \\
			\\
			\\
			(Every Inference has a maximum of 25 words)                                                                                                                                                                                                                                                                                                                                    \\
			(The reasoning process should be completed in multiple steps. The information in each step should be as concise as possible, and no redundant information should be included between each step.)                                                                                                                                                                               \\
			...                                                                                                                                                                                                                                                                                                                                                                            \\
			\bottomrule[1pt]
		\end{tabularx}
	}
	\caption{Prompt for Evidence Generation}
	\label{evidence prompt}
\end{table*}

\begin{table*}[h]
	\centering
	\resizebox{0.8\linewidth}{!}{
		\begin{tabularx}{15.5cm}{X}
			\toprule[1pt]
			I am an excellent linguist. Your task is to generate an opposite sentence.                              \\
			Given an original sentence, you should generate a new sentence to refute this sentence.                 \\
			Be sure to use creative methods instead of simply adding negative words.                                \\
			You will be provided with                                                                               \\
			\{\textbf{Original sentence}\} is the original sentence we provide;                                     \\
			\{\textbf{Opposite sentence}\} is the Opposite sentence, which will refute the original sentence.       \\
			Mandatory constraints: no unnecessary extra information can be induced.                                 \\
			\\
			\textbf{Example:}                                                                                       \\
			\textbf{Original sentence}: It was unlikely that any sequels would be made in the Power Rangers series. \\
			\textbf{Opposite sentence}: It's still possible that the Power Rangers series will get a sequel.        \\
			...                                                                                                     \\
			\#Just put the output after \{Opposite sentence\}. you are not allowed to do anything else.             \\
			\bottomrule[1pt]
		\end{tabularx}
	}
	\caption{Prompt for Counterfactual Effect Generation}
	\label{Counterfactual prompt}
\end{table*}

\begin{table*}[h]
	\centering
	\resizebox{0.8\linewidth}{!}{
		\begin{tabularx}{15.5cm}{X}
			\toprule[1pt]
			\textbf{class FACT}                                                          \\
			\textbf{<FACT CLASS PROMPT>}                                                 \\
			\# I am an excellent logician and programmer.                                \\
			\textbf{<TASK PROMPT> }                                                      \\
			\textbf{Example:}                                                            \\
			Claim = FACT("...")                                                          \\
			Sentence1 = FACT ("...")                                                     \\
			Sentence2 = FACT ("...")                                                     \\
			Sentence3 = FACT ("...")                                                     \\
			Sentence4 = FACT ("...")                                                     \\
			def build\_tree\_proof():                                                    \\
			\;\;\;\;one\_to\_multiple\_linking(Sentence2, [Sentence3, Sentence4])        \\
			\;\;\;\;multiple\_to\_one\_linking([Sentence1, Sentence3, Sentence4], Claim) \\
			...                                                                          \\
			\bottomrule[1pt]
		\end{tabularx}
	}
	\caption{Argument Structure Generation Prompt}
	\label{code prompt}
\end{table*}

\begin{figure*}[h]
	\centering
	\includegraphics[width=0.9\textwidth]{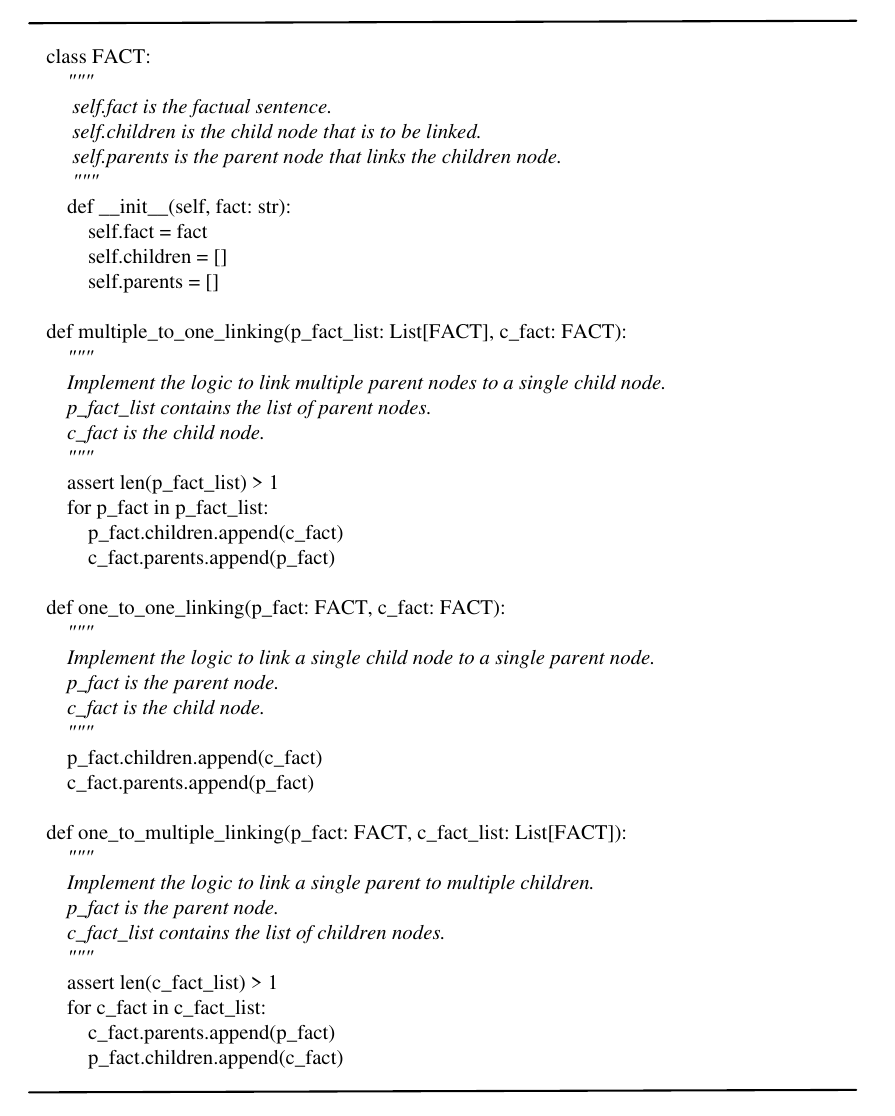}
	\caption{Argument Structure Generation Fact Class  Prompt}
	\label{FACT Prompt}
\end{figure*}

\begin{figure*}[h]
	\centering
	\includegraphics[width=0.9\textwidth]{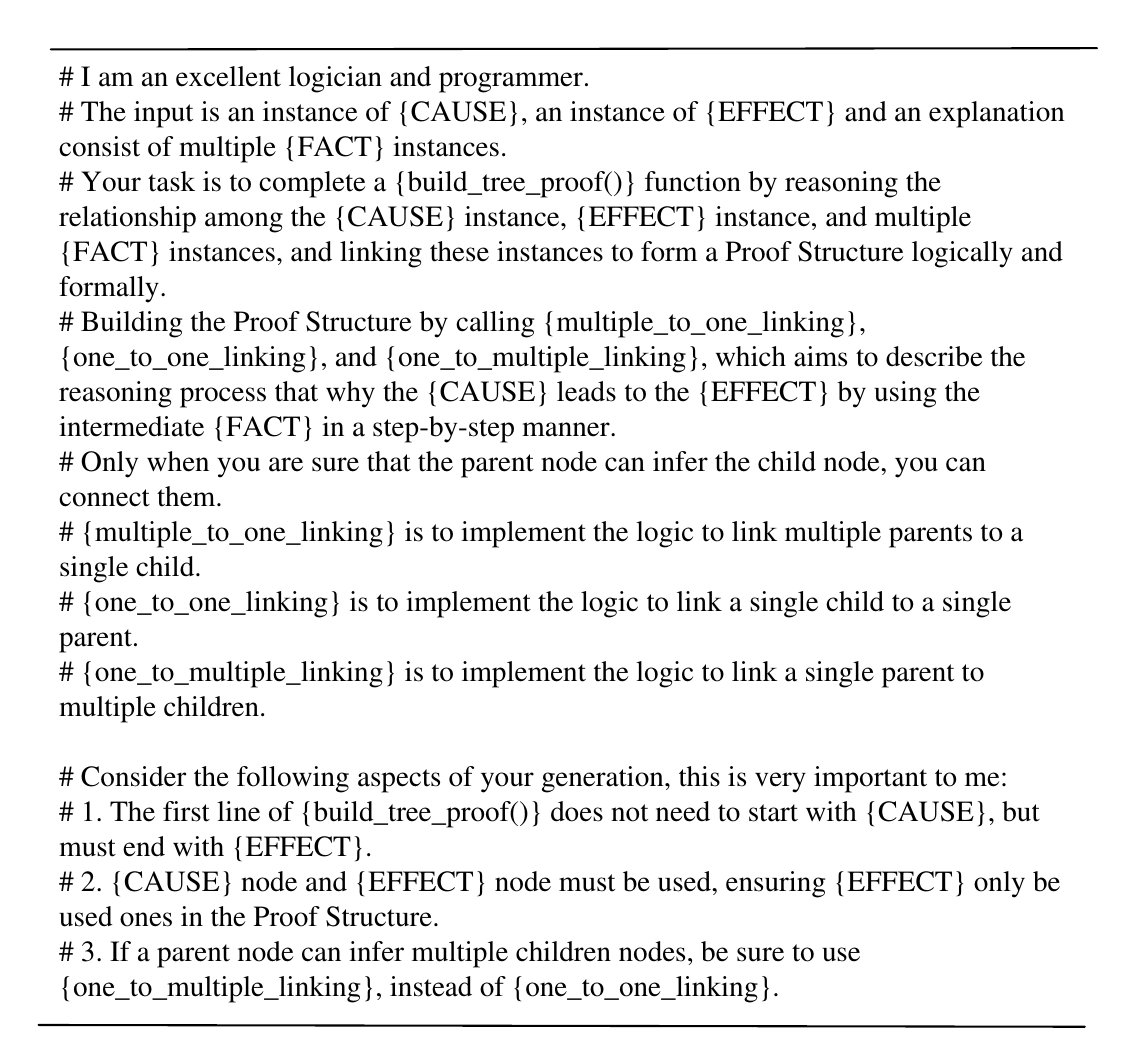}
	\caption{Argument Structure Generation Task Prompt Example}
	\label{Task Prompt}
\end{figure*}

\begin{table*}[h]
	\centering
	\resizebox{0.8\linewidth}{!}{
		\begin{tabularx}{15.5cm}{X}
			\toprule[1pt]

			I am an excellent logician and fact checker.                                                                                                                                                                     \\
			My task is to verify whether the claim is \{SUPPORTS\}, \{REFUTES\} or \{NOT ENOUGH INFO\}, based on \{Evidence\}.                                                                                               \\
			No need to focus on whether the given evidence is correct.                                                                                                                                                       \\
			(No need to return anything else superfluous)                                                                                                                                                                    \\
			\textbf{Example:}                                                                                                                                                                                                \\
			\textbf{\# Input:}                                                                                                                                                                                               \\
			\textbf{Claim}: Joanna Briscoe attributes the novel 'The Great Lover's evocative "sense of time and place" to its adept treatment of distinct elements such as Fabianism and class politics in that British era. \\
			\textbf{\# Evidence:}                                                                                                                                                                                            \\
			...                                                                                                                                                                                                              \\
			Determine whether \{Claim\} is supported or refuted:                                                                                                                                                             \\
			\textbf{\# Output:}                                                                                                                                                                                              \\
			\{SUPPORTS\}                                                                                                                                                                                                     \\
			\bottomrule[1pt]
		\end{tabularx}
	}
	\caption{Prompt for Task1}
	\label{task1 prompt}
\end{table*}

\begin{table*}[h]
	\centering
	\resizebox{0.8\linewidth}{!}{
		\begin{tabularx}{15.5cm}{X}
			\toprule[1pt]
			\textbf{<FACT CLASS PROMPT>}                                                          \\
			\\
			\textbf{<TASK PROMPT>}                                                                \\
			\\
			\textbf{Example:}                                                                     \\
			\textbf{\# Input:}                                                                    \\
			Claim = FACT("...")                                                                   \\
			\textbf{\# Evidence:}                                                                 \\
			Sentence1 = FACT("...")                                                               \\
			...                                                                                   \\
			\\
			def build\_tree\_proof():                                                             \\
			\;\;\;\;\;one\_to\_one\_linking(Sentence13, Sentence16)                               \\
			\;\;\;\;\;one\_to\_one\_linking(Sentence13, Sentence17)                               \\
			\;\;\;\;\;one\_to\_one\_linking(Sentence13, Sentence18)                               \\
			\;\;\;\;\;one\_to\_one\_linking(Sentence11, Sentence17)                               \\
			\;\;\;\;\;multiple\_to\_one\_linking({[}Sentence16, Sentence17, Sentence18{]}, Claim) \\
			\\
			\# Determine whether \{Claim\} is supported or refuted:                               \\
			\textbf{\# Output:}                                                                   \\
			print("REFUTES")                                                                      \\
			\bottomrule[1pt]
		\end{tabularx}
	}
	\caption{Prompt for Task2}
	\label{task2 prompt}
\end{table*}

\begin{table*}[h]
	\centering
	\resizebox{0.8\linewidth}{!}{
		\begin{tabularx}{15.5cm}{X}
			\toprule[1pt]
			\textbf{<FACT CLASS PROMPT>}                                                    \\
			\\
			\textbf{<TASK PROMPT>}                                                          \\
			\\
			\textbf{Example:}                                                               \\
			\textbf{\# Input: }                                                             \\
			Claim = FACT("...")                                                             \\
			\textbf{\# Evidence:}                                                           \\
			Sentence1 = FACT("...")                                                         \\
			...                                                                             \\
			\# Determine whether \{Claim\} is supported or refuted, complete this function: \\
			def build\_tree\_proof():                                                       \\
			\# \textbf{Output:}                                                             \\
			```python                                                                       \\
			one\_to\_one\_linking(Sentence8, Sentence15)                                    \\
			one\_to\_one\_linking(Sentence15, Sentence16)                                   \\
			one\_to\_one\_linking(Sentence16, Sentence17)                                   \\
			one\_to\_one\_linking(Sentence17, Sentence18)                                   \\
			one\_to\_one\_linking(Sentence18, Claim)                                        \\
			print("SUPPORTS")                                                               \\
			```                                                                             \\
			\bottomrule[1pt]
		\end{tabularx}
	}
	\caption{Prompt for Task3}
	\label{task3 prompt}
\end{table*}

\begin{table*}[h]
	\centering
	\resizebox{0.8\linewidth}{!}{
		\begin{tabularx}{15.5cm}{X}
			\toprule[1pt]
			\textbf{<FACT CLASS PROMPT>}                                                    \\
			\\
			\textbf{<TASK PROMPT>}                                                          \\
			\\
			\textbf{Example:}                                                               \\
			\textbf{\# Input: }                                                             \\
			Claim = FACT("...")                                                             \\
			\textbf{\# Evidence:}                                                           \\
			Sentence1 = FACT("...")                                                         \\
			...                                                                             \\
			\# Determine whether \{Claim\} is supported or refuted, complete this function: \\
			def build\_tree\_proof():                                                       \\
			\# \textbf{Output:}                                                             \\
			```python                                                                       \\
			Sentence15 = FACT("...")                                                        \\
			one\_to\_one\_linking(Sentence8, Sentence15)                                    \\
			Sentence16 = FACT("...")                                                        \\
			one\_to\_one\_linking(Sentence15, Sentence16)                                   \\
			Sentence17 = FACT("...")                                                        \\
			one\_to\_one\_linking(Sentence16, Sentence17)                                   \\
			Sentence18 = FACT("...")                                                        \\
			one\_to\_one\_linking(Sentence17, Sentence18)                                   \\
			one\_to\_one\_linking(Sentence18, Claim)                                        \\
			print("SUPPORTS")                                                               \\
			```                                                                             \\
			\bottomrule[1pt]
		\end{tabularx}
	}
	\caption{Prompt for Task4}
	\label{task4 prompt}
\end{table*}

\end{document}